\documentclass[11pt]{article}
\usepackage[margin=1in]{geometry}
\usepackage{setspace}
\usepackage{array}
\usepackage{url}
\usepackage{tabularx}
\usepackage{tabulary}
\usepackage{multirow}
\usepackage{multicol}
\usepackage{url}
\usepackage{graphicx}
\usepackage{booktabs}
\usepackage{listings}
\usepackage{xcolor}
\definecolor{linkcolor}{HTML}{003399}
\definecolor{iconcolorcvwarning}{HTML}{F5BB17}
\definecolor{iconcolorcvcrimson}{HTML}{990000}
\usepackage[hypertexnames=false]{hyperref}
\hypersetup{
    colorlinks=true,
    breaklinks=true,
    citecolor=linkcolor,
    linkcolor=linkcolor,
    filecolor=magenta,
    urlcolor=linkcolor,
}
\usepackage{cleveref}

\usepackage[section]{placeins}

\usepackage{etoc}
\etocsetnexttocdepth{subsubsection}

\usepackage{amssymb}
\usepackage{longtable}

\begin{document}

\title{Refinement of an Epilepsy Dictionary through Human Annotation of Health-related posts on Instagram}

\renewcommand{\thefootnote}{\fnsymbol{footnote}}

\author{
    Aehong Min$^{1}\footnotemark[2]$,
    Xuan Wang$^{2,4}\footnotemark[2]$,
    Rion Brattig Correia$^{3,4}$,
    Jordan Rozum$^{4}$, \\
    Wendy R. Miller$^{5}$, \&
    Luis M. Rocha$^{4,3,*}$
    \\
    \small $^{1}$ Donald Bren School of Information \& Computer Sciences, University of California, Irvine, CA, USA\\
    \small $^{2}$ Luddy School of Informatics, Computing \& Engineering, Indiana University, Bloomington, IN, USA \\
    \small $^{3}$ Instituto Gulbenkian de Ciência, Oeiras, Portugal \\
    \small $^{4}$ Dept. of Systems Science \& Industrial Engineering, Binghamton University, Binghamton, NY, USA. \\
    \small $^{5}$ School of Nursing, Indiana University, Indianapolis, IN, USA \\
    \small * \url{rocha@bighamton.edu}
}

\footnotetext[2]{These authors contributed equally to this work.}
\date{}

\maketitle

\doublespacing

\begin{abstract} %

\textbf{Objective |}
To (1) identify health-related terms used on social media posts that do not precisely match the health-related meaning of terms in a biomedical dictionary, (2) decide which terms need to be removed in order to improve the quality of the dictionary in the scope of biomedical text mining tasks, (3) evaluate the effect of removing imprecise terms on such tasks, and (4) discuss how human-centered annotation complements automated annotation in social media mining for biomedical purposes.

\textbf{Materials and Methods |}
We used a dictionary built from biomedical terminology extracted from various sources such as DrugBank, MedDRA, MedlinePlus, TCMGeneDIT, to tag more than 8 million \textit{Instagram}  posts by users who have mentioned an epilepsy-relevant drug at least once, between 2010 and early 2016.
A random sample of 1,771 posts with 2,947 term matches was evaluated by human annotators to identify false-positives. 
Frequent terms with a high false-positive rate were removed from the dictionary. To study the effect of removing those terms, we constructed knowledge networks using the refined and the original dictionaries and performed an eigenvector-centrality analysis on both networks.
OpenAI's GPT series models were compared against human annotation.

\textbf{Results |}

Analysis of the estimated false-positive rates of the annotated terms revealed 8 ambiguous terms (plus synonyms) used in \textit{Instagram}  posts, which were removed from the original dictionary. 
We show that the refined dictionary thus produced leads to a significantly different rank of important terms, as measured by their eigenvector-centrality of the knowledge networks.
Furthermore, the most important terms obtained after refinement are of greater medical relevance.
In addition, we show that OpenAI's GPT series models fare worse than human annotators in this task. 

\textbf{Discussion |}
Dictionaries built from traditional clinical terminology are not tailored for social media language and can bias results when used in biomedical inference pipelines, such as pharmacological surveillance. However, removing relatively few terms identified by human annotation significantly improves inference and recommendation of central terms and associations when studying social media cohorts.

\textbf{Conclusion |}
A human-centered methodology to refine biomedical dictionaries improves biomedical social media mining.

\end{abstract}

{

}

\section{Introduction}

\label{sec:intro}

Social media data, such as text, hashtags, or images in posts, allow researchers to gain unprecedented access to study human cohorts. It is now possible to quantitatively measure very large populations for their individual or collective experiences, behaviors, perceptions, and emotions \cite{Correia2020, guntuku2021social, klein2017detecting, Fan2017, saha2019social, Wood2022}.
Social media is more than a source of information or a valuable communication tool for its users \cite{Sultan2021}---it is also a valuable resource of information about human health and well-being \cite{Correia2020}.
It has been used to track and predict various public health issues \cite{kautz2013data}, such as mental health disorders \cite{Toulis2017, Lopez2020}, adverse drug reaction (ADR) \cite{Nikfarjam2015, Sarker:2015}, drug-drug interactions (DDI) \cite{yang2013, Hamed:2015, Correia2016}, and substance abuse \cite{sarker2016social, Garimella2016}.

\textit{Instagram} is one of the most popular social network services in the world; in 2021, \textit{Instagram} reached more than 1 billion users worldwide \cite{Enberg2020}.
Through an application for smartphones or tablets, users can share photos and videos, often accompanied by long captions.
Although most research on social media has been focused on data from \textit{Twitter}  (now named X) or \textit{Facebook} , \textit{Instagram}  has great potential for social media research given its increasing number of users.
It has already shown its potential for large-scale social media analysis and monitoring of public health issues, such as DDI and ADR, uncovering behavioral pathology and associations between drugs and symptoms in depression \cite{Correia2016}. 
It can even be used as a tool for communication and support between patients and healthcare providers \cite{Yakar2020}, and other health-related applications \cite{Correia2020}.
For these reasons, the work described here is focused on data from \textit{Instagram}, but our methodology and results are applicable to other social media from \textit{Twitter} to \textit{Reddit}  \cite{Guo23}.

Despite the benefits of social media analysis for public health, social media research has not focused much on people with epilepsy (PWE), a chronic noncommunicable brain disorder and one of the most common neurological diseases \cite{Hirtz2007}.
In the U.S., more than three million adults have epilepsy, and about 470,000 children were diagnosed with active epilepsy in 2015 \cite{Hirtz2007, WHO2019, Zack2017}.
Epilepsy-relevant \textit{Facebook}  pages and \textit{Twitter}  accounts play an essential role in providing information about drugs or correcting misconceptions or epilepsy stigma on online platforms \cite{McNeil2012, MENG201779}.
Furthermore, our team has shown that even small cohorts of epilepsy patients on \textit{Facebook}  can inform experts about relevant behaviors involved in rare outcomes, such as sudden death in epilepsy \cite{Wood2022}.
Therefore, more research on PWE and their caregivers' online behaviors on social media, from epilepsy-specific online groups to general-purpose platforms, is needed---especially to better understand their complex symptoms and medication schedules, including DDI and ADR.
To support such a research agenda, it is essential to develop automated annotation pipelines to mine and detect biomedical signals from large-scale social media data in general \cite{Correia2020, correia2024myaura}.
At the core of such pipelines is the construction of biomedical dictionaries to tag relevant terminology in social media posts. 
Typically these are produced from databases and named entity recognition tools that were developed for scientific discourse, such as papers with experimental evidence available on PubMed \cite{Correia2016, Correia2020, lourenco2011linear, zhang2022translational}.
However, it is unclear whether biomedical dictionaries built from scientific discourse and evidence are fit for the informal discourse and particularities of discussions on \textit{Instagram}  and other social media that are relevant for epilepsy (or other conditions of medical interest).

To address that question, here we present a human-centered dictionary refinement methodology and analysis tailored to tag clinically relevant terminology for the study of epilepsy cohorts on \textit{Instagram} .
For comparison, OpenAI's GPT series models, as representatives of Large Language Model (LLMs), were also used as an alternative annotation process, though leading  to worse results than human annotators in our analysis. 
In addition to producing a focused biomedical dictionary for social media, our manual annotation effort demonstrates that false positive terms (clinically relevant terms used in a clinically irrelevant context) with high frequency exist in social media discourse. 
In other words, dictionaries built from biomedical terminology appropriate for the scientific literature, contain terms that are used in social media with other meanings.
Moreover, those terms bias the knowledge inferences that automated pipelines might produce.
Indeed, we demonstrate that the removal of just a few high frequency false-positive dictionary terms improves the biomedical knowledge extracted from the epilepsy cohort on \textit{Instagram}, thus highlighting the importance of human annotation to improve the quality of cohort-specific social media analysis.

\section{Data and Methods}
\label{sec:methods}

\subsection{Dictionary Construction}

Our dictionary includes terms related to drugs, allergens, medical terms, and natural products, including cannabis. We construct this dictionary following \cite{Correia2016,correia2019PredictionDrugInteraction}. We recall the deails of that construction below.
We obtained these terms from a variety of existing medical ontologies and data sources.
Drug, allergen, and food terms are retrieved from \textit{Drugbank} (v.5.1.0) \cite{DrugBank5}.
Medical terms including, but not limited to, disease symptoms and drug side effects are obtained from \textit{MedDRA} (v.15) \cite{MedDRA}.
Natural products are retrieved from \textit{MedlinePlus} \cite{MedlinePlus} and \textit{TCMGeneDIT} \cite{TCMGeneDIT}.
For \texttt{cannabis}, we manually added commonly referred terms and slang, such as `\texttt{Mary Jane}' and `\texttt{420}', to our dictionary, as detailed in \cite{Correia2016}.
In addition, epilepsy terms commonly used by patients on Internet forums were manually added using a C-value \cite{Frantzi:2000} tokenizer on the \textit{Epilepsy.com} discussion forums.
These epilepsy-related terms include mentions such as `\texttt{VNS}' (i.e. Vagus Nerve Stimulator) and were validated by an epilepsy specialist and matched to \textit{MedDRA} codes.

We distinguish dictionary terms into four categories: Allergens, Drugs, Medical Terms, and Natural Products.
Allergens include food names, ingredients, and animals (e.g., \texttt{Orange, Duck});
Drugs include medicine and chemical compounds (e.g., \texttt{Diazepam});
Medical Terms include status and conditions of putative medical relevance, such as physical, psychological, or physiological features (e.g., \texttt{Headache, Feeling hot});
Natural Products consist of plants and their extracted elements (e.g., \texttt{Rose}).

Importantly, synonyms are possible for each term. Therefore, all those are matched---as child terms---to a unique \emph{parent} (\emph{preferred}) term.
For instance, \texttt{Weed}, \texttt{Mary jane}, \texttt{420} and \texttt{Cannabis} are all synonyms of the parent term \texttt{Cannabis}.
The parent term is also included as a child term for completeness of synonym lists.
Drug names are treated similarly, whereby we keep the chemical name as parent terms (e.g. \texttt{Diazepam}), and all known commercial names (as extracted from \textit{DrugBank} \cite{DrugBank5}) as child terms (e.g. \texttt{Valium}).
Some data sources of our dictionary already have term hierarchy, and we used it as the base of our parent term mapping (for example, the ``preferred term'' in MedDRA is mapped to the parent term in our dictionary). 
\Cref{tab:example} lists examples of the extracted posts, terms, and their parent term.

Drug brand names sometimes have very common names in the English language, which may increase the number of false-positive hits (e.g., \texttt{Nighttime} is a common word and a synonym for the drug \texttt{Benadryl}).
To account for that, we matched terms in our dictionary to the expected occurrence of such terms in the Brown Corpus \cite{BrownCorpus}.
Terms very commonly used in the daily English language were then ranked and removed.
After this collection and initial automatic curation procedure, our dictionary contains 176,278 terms, of which 105,345 are \textit{Drugs}, 66,961 are \textit{Medical Terms}, 2,797 are \textit{Allergens}, and 1,175 are \textit{Natural Products}.

\begin{table}[htbp]
    \centering
    \footnotesize
    \begin{tabular}{ p{2.2cm}|p{2.0cm}|p{2.2cm}|p{8cm} } 
    \toprule
    \textbf{Category} & \textbf{Frequency} & \textbf{Parent Term} & \textbf{Examples}\\
    \midrule

\multirow{3}{*}{Allergen} & \multirow{3}{*}{874 (29.7\%)} & Coffee bean & new nails, new bag and a \textcolor{red}{coffee} catch up makes for a very happy girl \#smiley \#girl \#pink [...] \#positivity \#goodday \#instacollage\\\cline{3-4}
     & & Cocoa & strawberry \textcolor{red}{chocolate} tini i made at work yesterday!! so adorable lol. [...]\\\cline{3-4}
     & & Tea leaf & studying with a cup of \textcolor{red}{tea} makes me feel british, bringing my own teabag just makes me cheap , wishing it was coffee . lol \#tea \#cheap \#wishfulthinking \#cuppatea\\
 \hline
\multirow{3}{*}{Drug} & \multirow{3}{*}{204 (6.9\%)} & Diazepam & saturday night survival kit \#goodbook \#warmbed \#hottea \#\textcolor{red}{valium} \#davidrakoff\\\cline{3-4}
     & & Ethanol & anyone got a number for a good rehab centre , think i'll be booking lottie in soon ! \#alcoholic \#\textcolor{red}{alcohol} \#friend \#crazy \#needshelp […]\\\cline{3-4}
     & & Caffeine & there's nothing better than a cuppa take-away coffee. except getting chocolate with your take away coffee! \#lifesgood! \#coffee \#\textcolor{red}{caffeine} \#java \#chocolate \#living […]\\
 \hline
\multirow{3}{*}{Medical Term} & \multirow{3}{*}{1647 (55.9\%)} & Feeling hot & \#day16 : hanging out in \#laos! and having this \#icecream \#dessert in this \textcolor{red}{hot} weather! \#wanderlust \#travel \#100happydays\\\cline{3-4}
     & & Tattoo & \#tattoolove. \#tattoohumour. \#tattooquote. \#\textcolor{red}{tattoo}. \#ilovetattoos. \#bodyart. \#insta. [...]\\\cline{3-4}
     & & Nasopharyngitis & i have a \textcolor{red}{cold}. any remedies? xx \#bed \#hottowelaroundyourhead \#sniff \#toiletpapertissues\\
 \hline
\multirow{3}{*}{Natural Product} & \multirow{3}{*}{222 (7.5\%)} & Cannabis & medicating made simple. \#vaping \#cannabidiol \#\textcolor{red}{cannabis} \#cbd \#monday \#itsgoingtobeagreatweek [...]\\\cline{3-4}
     & & Rose & […] \#my \#girl \#beauty \#\textcolor{red}{rose} \#roses \#flower \#flowers […]\\\cline{3-4}
     & & Rosa & […] \#primavera.  \#rose. \#instaflowers. \#\textcolor{red}{rosa}. \#absolutelyfabulous. \#roses. \#verde.\\
    \bottomrule
    \end{tabular}
    \caption{
        Examples of posts in the annotation sample with matched child terms (in red)  by category, frequency, and parent term.
    }
    \label{tab:example}
\end{table}

\subsection{Data Collection and Post Tagging}

Since June 2016, the collection of \textit{Instagram}  data via the platform's API has been limited due to a company policy change.
Here we use publicly available data from \textit{Instagram}  posts ranging from 2011 to early 2016 when they were collected and securely stored according to the platform's terms on our servers \cite{Correia2016}.
The data is comprised of the entire timelines (all time-stamped public posts) of Instagram users who produced at least one post mentioning a hashtag (i.e., \#) with a drug name (or synonym) known to treat epilepsy.
The (parent) drug names we used to retrieve timelines include \texttt{carbamazepine}, \texttt{clobazam}, \texttt{diazepam}, \texttt{lacosamide}, \texttt{lamotrigine}, \texttt{levetiracetam}, \texttt{oxcarbazepine}, as well as all their brand name (child term) synonyms (e.g., \texttt{Valium}).
In addition, we added all user timelines that mentioned the epilepsy-associated hashtag `\#seizuremeds' which is commonly used among PWE on discussion forums such as Epilepsy.com.
This resulted in the collection of the entire timelines of a cohort of 9,890 users, comprising 8,496,124 posts.
Duplicate posts from regrams were removed.
In order to protect user privacy, we did not extract demographic information from the collected accounts.

The caption field of all collected posts was subsequently tagged with the dictionary terms according to an automatic multi-word lexical matching pipeline, resulting in a total of 979,683 dictionary term matches on the more than 8 million Instagram posts on the dataset.

\section{Results}
\label{sec:results}

\subsection{Human-centered Annotation}
\label{sec:results-annotation}

Automatic lexical matching with biomedical dictionaries built from scientific discourse and evidence is not necessarily contextually accurate when tagging social media posts. 
Indeed, dictionary terms used in social media discourse may refer to alternate meanings without any putative clinical relevance.
For instance, the term `\texttt{hot}' has multiple meanings depending on context, ranging from `having a fever' to  `sexual appeal'.
Naturally, we are only interested in the potential clinical relevance of dictionary term usage.
Therefore, we need to refine the dictionary terms to improve the accuracy of term matches in biomedical social media analytics. That is, the goal is to reduce false positive term matches, such as in the example above.
To do this we designed and pursued a manual annotation workflow to identify the dictionary terms most likely to be false positives in the context of \textit{Instagram}  discourse related to epilepsy.

Because it was unfeasible for our team to manually annotate over 8 million posts, a sample was randomly selected to provide a reasonable amount of posts for human annotators. This resulted in a set of 1,771 posts containing at least one matched term, for a total of 2,947 matches, associated with 466 unique parent terms.
The number and proportion of matches per dictionary category is: $|Allergens|$ = 874 (29.7\%; 108 parent terms), $|Drugs|$ = 204 (6.9\%; 64 parent terms), $|Medical$ $terms|$ = 1,647 (55.9\%; 268 parent terms), and  $|Natural$ $products|$ = 222 (7.5\%; 26 parent terms).
\Cref{tab:example} shows examples of posts and respective matched child terms (in red), their parent terms and categories.

Each human annotator was given our annotation guidelines to understand the goal of the study and criteria to determine whether a matched term is used in the expected or correct sense (`true-positive') or not (`false-positive').
Also, they were provided with instructions with examples to learn how to annotate through our annotation tool.
This sample was subsequently used in our annotation workflow which comprises two rounds (See \Cref{fig:ant_process} \& SI: \Cref{si:ant_guide_exmpl}):

\begin{figure}
    \centering
    \includegraphics[width=0.9\linewidth]{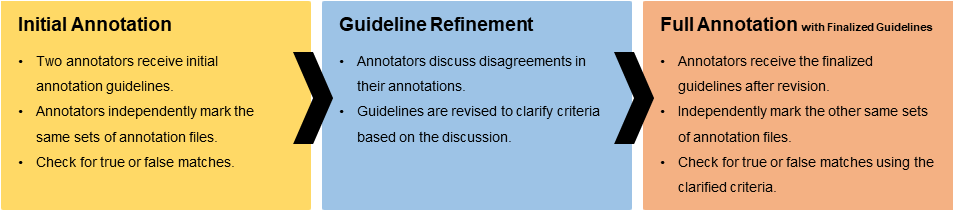}
    \caption{
        Manual annotation workflow
    }
    \label{fig:ant_process}
\end{figure}
\begin{enumerate}
    \item \textbf{Initial annotation \& guideline refinement}. 292 posts with 499 dictionary matches were used to test the annotation workflow and to refine our annotation guidelines, establishing a standard for deciding whether a matched term was used as intended by the biomedical dictionary (true positive) or not (false positive) in the context of the post. For instance, in the context of medical terms, if a matched term ``\texttt{A}'' is expressed as signifying a medical term, health condition, drug, food, or (potential) allergen (`A'), an annotator is instructed to label it as `True.' Conversely, if the term is expressed in a manner other than the intended dictionary meaning, but rather represents a metaphor, proper noun, or anything else, the annotator marks it as `False.' Each sampled post was assigned to two annotators, with disagreements being collectively discussed to reach a consensus that triggered a revision to the guidelines.
 \item \textbf{Full annotation review}. Using the refined annotation guidelines obtained from the first step of the workflow (See \Cref{fig:ant_guide_exmpl}), all 1,771 posts in the sample and their 2,947 matches were reviewed independently by two annotators.
Data scientists in training as well as epilepsy researchers participated in the annotation.
They decided whether terms were \emph{appropriately matched} (true-positive), \emph{inappropriately matched} (false-positive), or \emph{unclear to determine} (e.g., a term with unclear meaning).
Annotations were compared, achieving a good inter-rater reliability rating (Cohen’s Kappa=0.634) \cite{mchugh2012interrater}.

\end{enumerate}

\begin{table}
    \centering
    \includegraphics[width=0.9\linewidth]{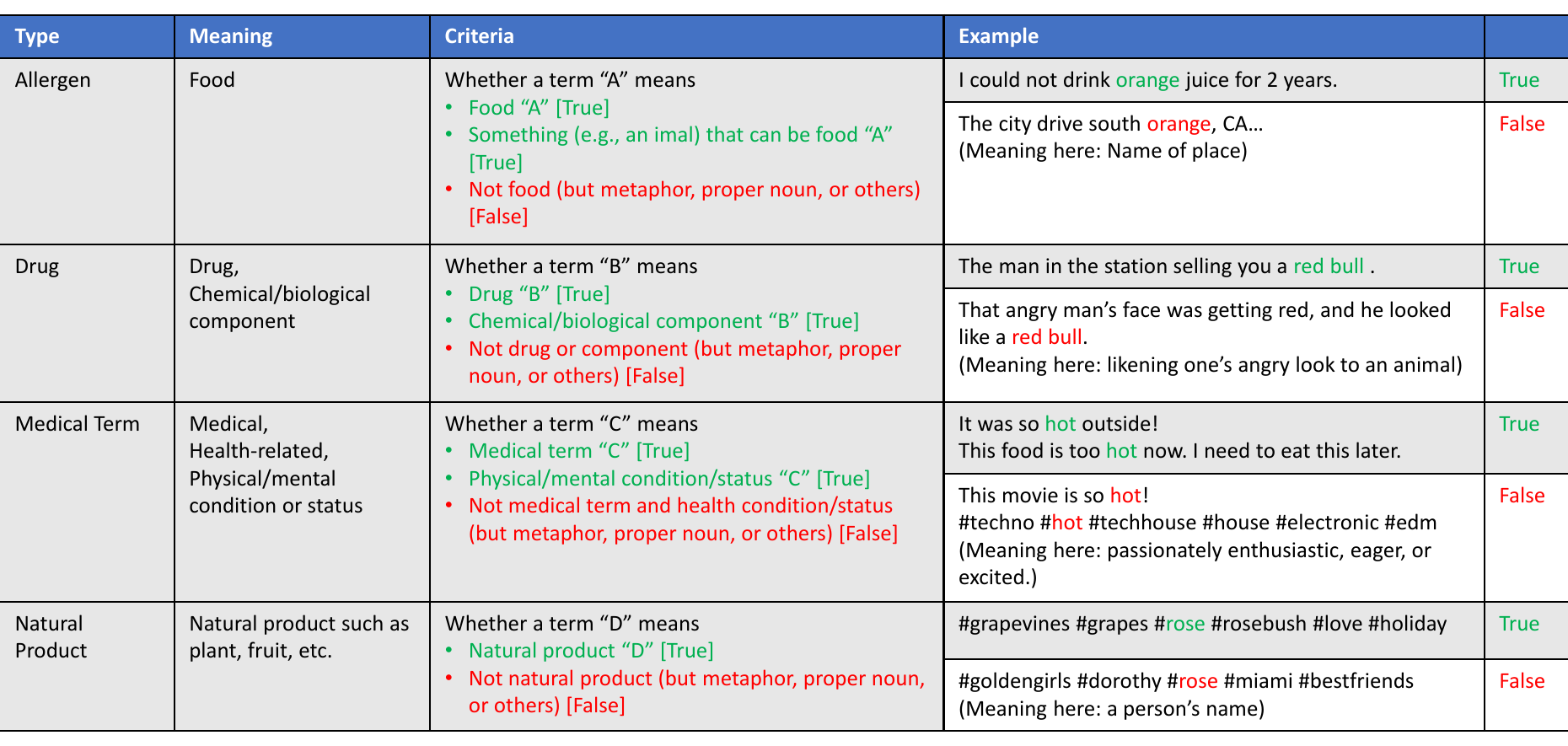}
    \caption{
        Finalized annotation guidelines: category \& criteria
    }
    \label{fig:ant_guide_exmpl}
\end{table}

An example of the interface used by annotators is provided in \Cref{fig:annotation} in SI. In this interface, each row displays a complete \textit{Instagram}  post with a dictionary term match highlighted in red. An annotator can review the context of each post and determine whether the meaning of the red-highlighted word is expressed with the intended use of the biomedical dictionary term, following the criteria provided in the guidelines.
Notice that annotators are not shown the photos that accompany the post text on Instagram. They are only shown the text, which makes this a difficult task for human annotators and more so for LLMs used below.

\subsection{Identifying Ambiguous Terms}

\begin{table}[htbp]
    \centering
    \footnotesize
    \begin{tabular}{ p{2.2cm}|p{1.8cm}|p{1.8cm}|p{8cm} } 
\toprule
\textbf{Category} & \textbf{Freq. of FP} & \textbf{Examples} & \\
\midrule
Allergen & 166 (19.0\%) & Orange & […] here our bottoms are shown in  \textcolor{red}{orange} hibiscus and neon orange paired with coordinating mahina tops. \# aquabikini2016 \#summer \#bikini \#etsy \#handmadebikini […] \\
\hline
Drug & 35 (17.2\%) & Diazepam & quiz team champions of the world \#winners \#\textcolor{red}{valium} \#kimye \# getyournotesoutforthelads \#mensfashion \#vscocam \#vsco. \#desree. \#life. \#noghost. \#pieceoftoast.\\
\hline
Medical Term & 571 (34.7\%) & Feeling hot & my jam \#tool \#adjustor \#dj \#kay d \#smith \#crazy \#\textcolor{red}{hot} \#techhouse \#hard \#techno \#house \#electronic \#edm \#edmlife \#realedm \#dance \#plur \#rave \#drums \#kick \#synth \#snare \#epic\\
\hline
Natural Product & 67 (30.2\%) & Rose & as its my birthday tomorrow, wine not!? \#itsmybirthdaytomorrow \#twentythree \#gallo \#zinfandel \#\textcolor{red}{rose} \#wine \#black \#vapour \#electriccigerette \#cherrymenthol\\
\bottomrule
    \end{tabular}
    \caption{
        Examples of false-positive terms, and overall false-positive rate per term category.
    }
    \label{tab:example-fp}
\end{table}

After the full annotation review step of the human-centered annotation process,  839 (28.5\%) false-positive term matches were observed---by annotators considering and inferring the context of the post where the match occurred from text alone. Examples of posts with false-positive matches are shown in Table \ref{tab:example-fp}. 
In one case (top row), from context, annotators infer that the dictionary term for \texttt{Orange} (as fruit it is in the
allergen
category) is actually used to mean the color orange.
Similarly, in another post (bottom row), \texttt{rose} was actually identified as a typo for \texttt{rosé} (wine), thus assigned to an incorrect parent dictionary term.

Focusing on each of the four term categories, the human-centered annotation revealed the following false-positive rates (including unclear cases):

\begin{itemize}
    \item \textit{Allergen}: 166 out of 874 (19\%). 
These include \texttt{Orange}, \texttt{Apple} and \texttt{Ginger} which were frequently found to indicate a color, name (brand, pet, place, etc.), or toys, respectively.
     \item \textit{Drug}: 35 out of 204 (17\%). 
The match for \texttt{Valium} (brand name for \texttt{diazepam}) had the highest false positive rate in this category. 
Since this drug brand name is frequently used as a metaphor on social media (unlike in the scientific literature), a discourse feature human annotators can easily infer (unlike most automatic methods).
\item \textit{Medical term}: 571 out of 1,647 (35\%). This is the category with the largest observed false-positive rate (including unclear cases), due to the possible broad meanings in which these terms were frequently used.
For instance, \texttt{Hot} resolves to the \texttt{Feeling hot} parent term, which clinically means to be ``having or feeling a relatively high temperature'' (fever).
However, in the sampled posts, \texttt{Hot} often meant ``sexually excited'' or ``newly made'' \cite{Merriam-WebsterInc.2020.hot}---another case of very distinct usage between social media and scientific discourse.
\item \textit{Natural product}: 67 out of 222 (30\%).

Examples include \texttt{Rose}, \texttt{Rosa}, and \texttt{Daphne}, where the inferred context often indicated one's name or color.

\end{itemize}

\begin{figure}
    \centering
    \includegraphics{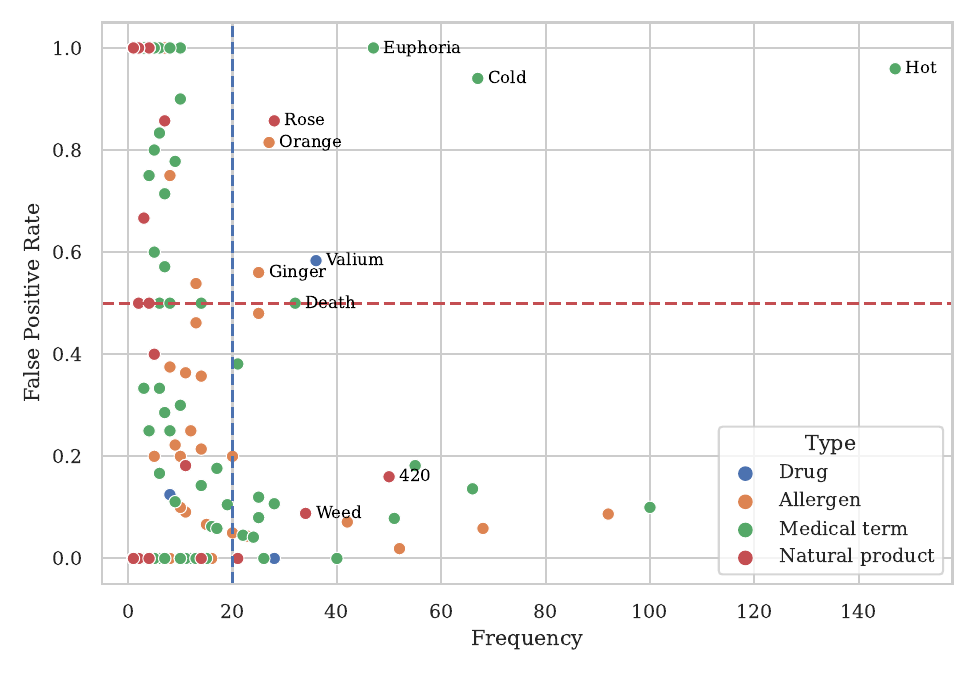}
    \caption{
        False-positive rate \& frequency of parent terms in the annotated sample of posts. The dashed horizontal line depicts a false-positive rate of 50\%.
    }
    \label{fig:chart1}
\end{figure}

Given the false-positive rates observed for term matches of the post sample, the next step is to identify which terms are most ambiguous and should be removed from the dictionary in order to refine and tailor it to epilepsy discourse on social media, our problem domain.
\Cref{fig:chart1} charts the false-positive rate against the frequency of all parent terms evaluated by the manual annotation workflow described above.
The most concerning terms are naturally those that occur very frequently in our data and simultaneously have a high false-positive rate---those on the top and right of \Cref{fig:chart1}.
\Cref{tab:top8} lists the top 8 such child terms, as well as their respective parent terms.
For instance, the terms \texttt{Hot}, \texttt{Cold} and \texttt{Euphoria} are frequent matches in posts, but most of the time without any putative medical significance with false-positive rates above 90\%, since they typically occur in contexts without any relationship to the intended meaning of the term in the biomedical dictionary.
The same can be said for the terms \texttt{Rose} and \texttt{Orange}, though less frequently  (fewer than 30 matches) but still unacceptably high false-positive rates above 80\%.
In contrast, while the term \texttt{Cannabis} appears very frequently in our samples (144 total matches for all of its child terms), it is most often used in relevant contexts (126 matches (88\%)). In \Cref{fig:chart1}, we show two of the synonyms of \texttt{Cannabis}: \texttt{420} and \texttt{weed}.

\begin{table}[htbp]
    \centering
    \footnotesize
    \begin{tabulary}{\linewidth}{ c|l|L|l|r|C } 
    \toprule
\textbf{Rank} & \textbf{Term} & \textbf{Parent Term} & \textbf{Category} & \textbf{Frequency} & \textbf{False-positive Rate}\\
\midrule
1 & Hot & Feeling hot & Medical term & 147 & 0.96\\[0pt]
2 & Cold & Nasopharyngitis & Medical term & 67 & 0.94\\[0pt]
3 & Euphoria & Euphoric mood & Medical term & 47 & 1.00\\[0pt]
4 & Valium & Diazepam & Drug & 36 & 0.58\\[0pt]
5 & Death & Death & Medical term & 32 & 0.50\\[0pt]
6 & Rose & Rose & Natural product & 28 & 0.86\\[0pt]
7 & Orange & Orange & Allergen & 27 & 0.81\\[0pt]
8 & Ginger & Ginger & Allergen & 25 & 0.56\\[0pt]
    \bottomrule
    \end{tabulary}
    \caption{
        Top 8 terms with the largest frequencies and false-positive rates.
    }
    \label{tab:top8}
\end{table}

To obtain a new \emph{refined dictionary} we established a criterion to remove terms that maximize false-positive rate and occur frequently enough in the human-annotated post sample.
We first selected terms with false-positive rates $\geq 0.5$ and frequency $\geq 20$ (red and blue dashed lines in \Cref{fig:chart1}, respectively), subsequently ranking them by frequency.
This resulted in the removal of 8 terms listed in \Cref{tab:top8}:
 \texttt{Hot}, \texttt{Cold}, \texttt{Euphoria}, \texttt{Valium}, \texttt{Death}, \texttt{Rose}, \texttt{Orange}, and \texttt{Ginger}.
Among the 8 terms, we found that most occurrences of \texttt{Hot}  were not about feeling elevated temperature but about feeling excitement or describing something as popular. On the contrary, \texttt{Cold} was mostly used as a term about temperature, not about the illness known as the Common Cold. Likewise, the meaning of \texttt{Euphoria} was often not related to one's feeling of joy, since it was frequently used in club music-relevant contexts that may indicate a musical genre, a music album, or a brand. As for \texttt{Valium}, if the word was mentioned with other terms related to drugs, it was easier to determine whether it is a true- or false-positive term. However, almost half of the posts that mentioned it did not provide enough context. In the remaining half, it was often used as a metaphor for something boring (e.g. example on the second row of \cref{tab:example-fp}). Finally, \texttt{Rose}, \texttt{Orange}, and \texttt{Ginger} in the posts were frequently used as proper nouns, such as a person, an area, or a pet. 

Notice that we could have removed more than these 8 terms with a false-positive rate above 50\% (red dashed line in \Cref{fig:chart1}). 
However, such terms have low frequencies of matches in our sample, which means that the false-positive rate was estimated with few human-annotated matches and potentially subjected to fluctuations from small sample size. 
Thus, to be conservative given that terms originate from established medical dictionaries, we retained them. 
In any case, as we show next, the impact of removing these 8 is much larger than removing other terms with similar frequency.

\subsection{Impact of Removing Ambiguous Terms}

After dictionary refinement, we evaluate the impact of term removal on the eigenvector-centrality 
of networks whose edge weights are obtained by tallying dictionary co-mentions in posts (their construction, especially how we dealt with parent terms and child terms, is detailed in SI \Cref{si:sec:comention}).
These co-mention networks can be seen as associative knowledge structures that characterize the discourse of social media cohorts \cite{hamed2015twitter, simas2015distance,Correia2016,correia2024myaura}. 

The co-mention networks is built on the level of parent terms in the sense that each node is a parent term covering several synonyms in the dictionary.

In such associative knowledge networks, removing terms from the dictionary potentially reduces the set of nodes that comprise them (each parent term is represented as a node and the term removal is done in the level of child term). More importantly, removing terms also affects the edge weights between nodes, as co-mention counts are altered. Since the structure of connections is altered, all inferences one can make from these networks---such as information retrieval, link prediction, community structure, shortest paths \cite{simas2015distance}---can be affected.

We chose eigenvector centrality because it is a popular measurement of node importance that accounts for indirect influence between nodes on undirected graphs, thus allowing us to assess the network-level impact of term removal \cite{Bonacich:1986}. 
Another popular node centrality is PageRank centrality. Although it is powerful on directed graphs, it is highly correlated with (or almost exactly the same as) node degree when applied to undirected networks \cite{perraSpectralCentralityMeasures2008, grolmuszNotePageRankUndirected2015}. Therefore, since our co-mention networks are undirected,  we choose eigenvector centrality rather than PageRank to assess the effect of removing terms.

\begin{table}[htbp]
    \centering
    \footnotesize
    \begin{tabular}{rlclc} 
    \toprule
& \multicolumn{2}{c}{Original} & \multicolumn{2}{c}{Refined} \\
\midrule
Rank & Parent term & Centrality & Parent term & Centrality \\
\midrule
1 & Feeling hot & 0.701 & Depression & 0.599\\[0pt]
2 & Euphoric mood & 0.685 & Anxiety & 0.540\\[0pt]
3 & Cocoa & 0.087 & Completed suicide & 0.291\\[0pt]
4 & Nasopharyngitis & 0.064 & Pain & 0.230\\[0pt]
5 & Homosexuality & 0.060 & Insomnia & 0.218\\[0pt]
6 & Coffee bean & 0.059 & Decreased appetite & 0.165\\[0pt]
7 & Tattoo & 0.052 & Bulimia nervosa & 0.101\\[0pt]
8 & Tea leaf & 0.040 & Schizophrenia & 0.087\\[0pt]
9 & Depression & 0.040 & Agoraphobia & 0.086\\[0pt]
10 & Anxiety & 0.038 & Cannabis & 0.084\\[0pt]
11 & Vegan & 0.036 & Migraine & 0.081\\[0pt]
12 & Cannabis & 0.032 & Stress & 0.078\\[0pt]
13 & Pain & 0.032 & Weight & 0.069\\[0pt]
14 & Chicken & 0.026 & Psychotic disorder & 0.068\\[0pt]
15 & Orange & 0.021 & Vegan & 0.066\\[0pt]
16 & Death & 0.021 & Mania & 0.064\\[0pt]
17 & Hyperhidrosis & 0.020 & Fear & 0.061\\[0pt]
18 & Caffeine & 0.019 & Fibromyalgia & 0.057\\[0pt]
19 & Banana & 0.017 & Convulsion & 0.056\\[0pt]
20 & Lemon & 0.016 & Paranoia & 0.054\\[0pt]
    \bottomrule
    \end{tabular}
    \caption{
        Top 20 highest eigenvector centrality terms of the epilepsy cohort \textit{Instagram}  co-mention knowledge networks built with the original (left) and the refined (right) dictionary.
    }
    \label{tab:co-mention}
\end{table}

The top 20 terms ranked by eigenvector centrality, before and after dictionary refinement, are shown in \cref{tab:co-mention}. 
Top eigenvector centrality terms before the refinement include terms not particularly relevant to epilepsy, such as \texttt{Cocoa} or \texttt{Tattoo}.
However, after dictionary refinement, not onlydo parent terms like \texttt{Feeling hot} disappear from the top, since their main child terms were no longer present in the dictionary, but we see that all the top 10 terms are associated with epilepsy, as attested by our epilepsy specialists.
For instance, \texttt{Depression}, a clinical diagnosis often co-morbid with epilepsy, jumps from 9th (0.040) to the top ranked centrality score term (0.599), closely followed by \texttt{Anxiety}.
This suggests that our dictionary refinement improved the quality of the top eigenvector centrality terms, bringing epilepsy-related terms to a more central role in the knowledge network. 

\begin{figure}
    \centering
    \includegraphics{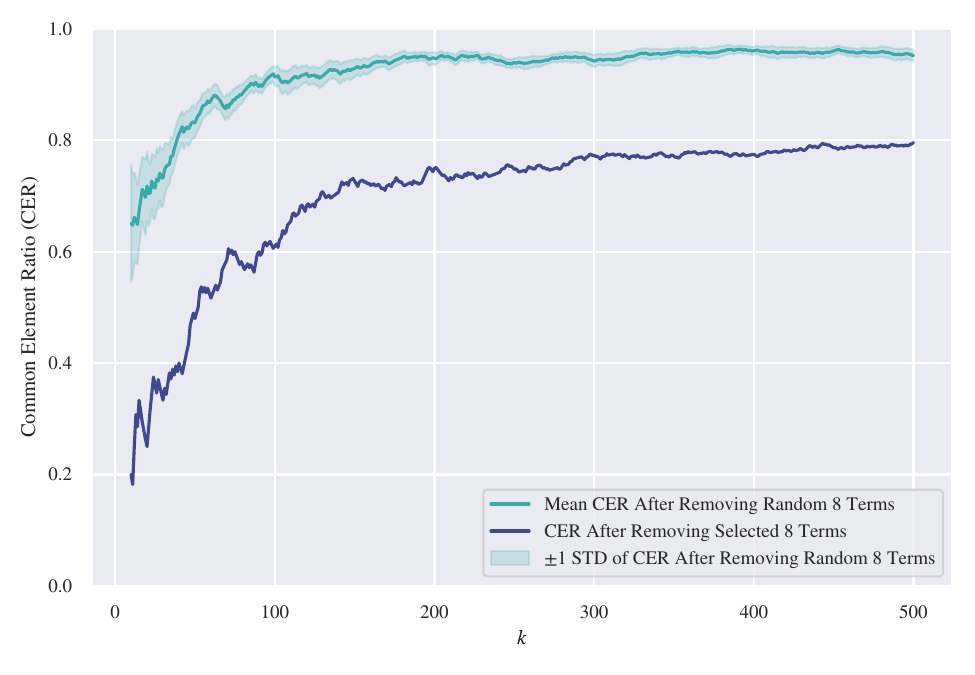}
    \caption{
        Impact of removing the selected 8 terms on the top $k$ ($10 \leq k \leq 500$) highest eigenvector centrality terms by calculating the common elements ratio between the top $k$ term lists before and after term removal. For example, if after term removal, 5 terms are still in the top 10 terms lists, the common element ratio will be $0.5$ for $k=10$. 
        The top $k$ eigenvector centrality terms lists were calculated on the \textit{Instagram}  co-mention network with or without term removal. 
        For baseline comparison, we generated 1,000 samples ($N=8$) from terms with a False Positive Rate less than 0.5 and a frequency greater than or equal to the minimum frequency of the 8 selected terms in our \textit{Instagram} data corpus, 10,230. 
    }
    \label{fig:null-selected}
\end{figure}

\begin{table}[htbp]
    \centering
    \footnotesize
    \begin{tabular}{cccc}
    \toprule

\(k\) & $K_{refined}$ & $K_{random}$  & $P_{obs}(K_{random}\geq K_{refined})$\\
\midrule
\phantom{0}10 & \phantom{000}108 & \phantom{000}48.1\phantom{0}$\pm$ \phantom{00}16.0 & 0.138 \\
\phantom{0}20 & \phantom{000}393 & \phantom{00}162.1\phantom{0}$\pm$ \phantom{00}40.5 & 0.000 \\
\phantom{0}50 & \phantom{00}1734 & \phantom{00}613.0\phantom{0}$\pm$ \phantom{00}84.8 & 0.000 \\
          100 & \phantom{00}6134 & \phantom{0}1528.6\phantom{0}$\pm$ \phantom{0}381.6 & 0.000 \\
          200 & \phantom{0}19021 & \phantom{0}4420.1\phantom{0}$\pm$           1153.6 & 0.000 \\
          500 &           101160 &           27797.9\phantom{0}$\pm$           6336.6 & 0.000 \\

\bottomrule
    \end{tabular}
    \caption{
        Fagin's generalization of Kendall's distance $K$ for top $k$ terms ranked by co-mention eigenvector centrality. Distances in the second column ($K_{refined}$) are computed between term rankings obtained using the original dictionary and those obtained using the refined dictionary that removes 8 frequent terms with a high false positive rate. The third column ($K_{random}$) shows the average distance (plus or minus one standard deviation) between the original ranking and the one obtained after removing an 8 randomly selected highly frequent terms with a false positive rate below $50\%$ (1,000 samples). The final column indicates the observed proportion of these samples that have a larger distance from the original ranking than the ranking obtained using our refinement.}
    
    \label{tab:fagin-tau}
\end{table}

To investigate whether the observed changes on the top eigenvector centrality terms are simply due the high frequency of the 8 removed terms, we computed a null model experiment in which 8 terms with similar frequencies but smaller false positive rates (below $50\%$) are randomly sampled 1,000 times and removed, with subsequent computation of eigenvector centrality of the resulting networks.
Because the top $k$ lists before and after dictionary refinement do not have the same set of terms, common rank difference metrics, such as Spearman's coefficient and Kendall's $\tau$ cannot be directly applied. 
In \Cref{fig:null-selected}, we compare the top $k$ terms by eigenvector centrality (of the dictionary term co-mention networks), before and after removing 8 dictionary terms, by counting the common elements ratio (CER) between the top $k$ lists for each case. 
It is very clear that the impact of removing the 8 terms with false positive rate above $50\%$ (via human-annotation) is much higher than when we remove 8 random terms with same or higher frequency, but lower false-positive rate.
For $k=10$ (the top 10 with largest eigenvector centrality), only $20\%$ of the terms remain after removal of the 8 terms with largest false positive rate per human annotation. Whereas, in the case of random removal of 8 terms with similar frequency, on average, $65\%$ of the terms remain unaffected in the top 10. 
The difference between the selected terms and the null model remains large for all values of $k$, even if CER increases.
While the impact of removing random terms almost saturates after $k \geq 100$, at 91\% CER, the impact of removing the 8 terms with high false positive rate remains significant for all $k$, with over 20\% different terms even for $k=500$.

Additionally, we also use Fagin's generalized Kendall's distance $K$, a metric specially designed to compare top $k$ ranked lists \cite{faginComparingTopLists2003}. We use it to quantify the impact on rank of removing the 8 ambiguous terms by computing the distance between the eigenvector centrality rankings obtained before and after removal. 
$K$ is comparable only to lists with the same sizes (the same $k$), with a higher value denoting a higher rank difference, with the penalty of missing elements considered. 
The value of $K$ obtained using our refined dictionary (denoted $K_{refined}$) for various values of $k$ is given in \Cref{tab:fagin-tau}, alongside the average distance obtained by removing 8 similarly frequent random terms (denoted $K_{random}$).
\Cref{tab:fagin-tau} shows that the 8 ambiguous terms selected through human annotation have statistically significantly larger impact on eigenvector centrality rankings (higher $K$) beyond the top 20 nodes than removal of similar random sets of terms. Though removing random high frequency terms could have a similar impact on the top 10 terms list in relatively rare cases (about 14\% of the time), we observed no case (out of 1,000 samples) in which this occurred for the top 20 terms or beyond. 
Moreover, the difference between $K$ for the 8 ambiguous terms and the average $K$ for the randomly removed terms increases dramatically as the list size $k$ increases.

The null model comparison, with both CER and Fagin’s generalized Kendall’s distance for rank comparison, demonstrates that removal of terms revealed as ambiguous by human annotation, have much higher impact on the knowledge network of biomedical terms than removing random terms with the same frequency (but lower false positive rate).
In other words, the high false-positive rate terms, estimated per human annotation, are likely confusing the co-mention network with many spurious edges that are not coherent with the remaining biomedical knowledge captured by the co-mention networks. Thus, removing them has a high impact on the network because their co-mention associations are quite different from those of other terms in dictionary.
Indeed, random removals of terms with similarly high frequency are not as impactful, because many of the associations the random terms induce are more coherent with the structure of other biomedical term associations.

In summary, the network global impact of removing human annotated high false-positive rate terms highlights the value of human context-specific annotation.

\subsection{Comparing Social Media with Medical and Scientific Discourse}

At the onset, the hypothesis behind our study is that while there is much discourse on general-purpose social media platforms that is of biomedical relevance, it is expressed differently than scientific discourse and unfolds simultaneously with many other contexts that are not relevant to health. Thus, human-centered dictionary refinement is needed to remove ambiguous terms in such platforms.

To emphasize this need, and how different general-purpose social media discourse is, we go beyond \textit{Instagram} and deploy both the original and the refined version of our dictionary in other data sources: epilepsy-related abstracts from \textit{PubMed}, epilepsy-related clinical trials from \href{https://clinicaltrials.gov}{clinicaltrials.gov}, \textit{Twitter} (now named X) posts from users who have mentioned epilepsy related terms, and discussion forums from \href{https://Epilepsy.com}{Epilepsy.com}. Detailed descriptions of data harvesting and network construction of additional data sources are available in SI \Cref{sec:si:refinement-effects} under each data source section. 
The first two additional data sources pertain to scientific discourse. 
Both \textit{Instagram} and \textit{Twitter} data sources represent epilepsy related users' speech on a general-purpose social media. The last data source is a form of social media that is focused on a single topic of medical relevance---epilepsy, in this case.

\begin{table}[htbp]
\caption{\label{tab:data-sources-comparison} The comparison of the impact of removing the 8 high false-positive rate high frequency term on different data sources, measured by Fagin's generalization of Kendall's distance $K$ for top $k$ terms ranked by eigenvector centrality. %
In parentheses, we divide the distances in the right four columns by the distances in the second column to show the reader how great the impact relative to the impact we observe on \textit{Instagram}. 
Distances are computed between term rankings obtained using the co-mention network constructed from the original dictionary and those obtained from the co-mention network constructed after dictionary refinement. EFF is short for the discussion forums from \href{https://Epilepsy.com}{Epilepsy.com}. CT is short for \href{https://clinicaltrials.gov}{clinicaltrials.gov}. }
\centering
\begin{tabular}{rrrrrr}
\toprule
\(k\) & Instagram & Twitter & EFF & PubMed & CT\\[0pt]
\hline

10 & 108 & 34(0.31) & 0(0.00) & 3(0.03) & 0(0.00)\\[0pt]
20 & 393 & 300(0.76) & 0(0.00) & 82(0.21) & 0(0.00)\\[0pt]
50 & 1734 & 974(0.56) & 101(0.06) & 300(0.17) & 82(0.05)\\[0pt]
100 & 6134 & 2418(0.39) & 344(0.06) & 1324(0.22) & 137(0.02)\\[0pt]
200 & 19021 & 8607(0.45) & 947(0.05) & 3816(0.20) & 270(0.01)\\[0pt]
500 & 101160 & 46050(0.46) & 1099(0.01) & 19594(0.19) & 1715(0.02)\\[0pt]

\bottomrule
\end{tabular}
\end{table}

In \Cref{tab:data-sources-comparison} we compare the impact of removing the 8 high false positive terms using Fagin's generalized Kendall's distance as in \Cref{tab:fagin-tau}. 
\textit{Twitter}, being a general purpose social media like \textit{Instagram}, received an impact approximately a half of \textit{Instagram} received after terms removal, as measured by Fagin's generalized Kendall's distance. 
Other data sources received much less impact. Networks built from \href{https://clinicaltrials.gov}{clinicaltrials.gov} data and the \href{https://Epilepsy.com}{Epilepsy.com} forums data set received zero impact on top 20 terms list and at most 6\% of the impact as \textit{Instagram} on other lists. PubMed, being a much larger text corpus, received only at most 22\% of the impact as \textit{Instagram}. 
This is also supported by \Cref{tab:si:twitter-eigen}, \Cref{tab:si:eff-eigen}, \Cref{tab:si:pubmed-eigen}, and \Cref{tab:si:ct-eigen} in SI, in which we show that aside from \textit{Twitter}, for all other data sources, there are only inconsequential impacts on the top 20 terms ranked by eigenvector centrality---except for the disappearance of those 8 terms we removed.
In other words, unlike in \textit{Instagram} and \textit{Twitter} , those 8 terms are either not ambiguous at all or their ambiguous usages do not affect the network analysis greatly in scientific database (\textit{PubMed} and \textit{Clinical trials}) nor in epilepsy-specific social media (\textit{Epilepsy Foundation Forums}) discourse.

This suggests that general-purpose social media platforms such as \textit{Instagram} are much noisier for biomedical surveillance, and for epilepsy research in particular, since users tend to post about many distinct aspects of their lives.
In contrast, in the Epilepsy Foundation (EF) forums users center discourse around their condition, which naturally makes it a rich resource for epilepsy research.
The same can be said for the chosen \textit{PubMed} articles or \textit{Clinical Trials}, where the scientific context is focused on epilepsy.
Therefore, the impact of removing the same 8 terms from the knowledge networks of both the EF forums and the \textit{PubMed} abstracts is much less pronounced in comparison to \textit{Instagram} .

\subsection{The disagreement between GPT-4 and human annotators is significant}
\label{sec:result-gpt}

Complementary to our human annotation efforts, we tested the feasibility of using a large language model instead of human annotators in the labeling process of our proposed dictionary refinement workflow. We assigned the same labeling task to OpenAI's latest and most advanced model, GPT-4 (version 1106 in 2023) and its precesssor GPT-3.5 via OpenAI's API, using the same guidelines we provided to human annotators \cite{brownLanguageModelsAre2020, openaiGPT4TechnicalReport2024}. 
We find significant disagreement between GPT-4 and the decisions of human annotators. 

We use OpenAI's API to ask OpenAI's GPT-3.5 (version gpt-3.5-turbo-1106) and GPT-4 (version gpt-4-1106-preview) to do the same labeling task as our human annotators. 
We converted the human annotation guidelines (detailed in \cref{sec:results-annotation} and summarized in \cref{fig:annotation}), initially designed for humans and presented to them in slides, into a text-based prompt for the AI, undergoing several rounds of refinement for optimization. 
For human annotators, the term matches to be annotated are highlighted in color, while for the LLM's task, these terms are marked with asterisks (\texttt{*}) on both sides. 
This method serves as a text-based alternative to visual highlighting,  and is commonly used in prompting LLMs. 

For each term tagged on a post, we query the LLM using two prompts: one system prompt based on the annotation guidelines, and one user prompt containing the entire text of post with the highlighted term. 
The LLM was instructed to provide a response in json format, with both the decision label and justification for the classification. 
The model's response was then collected and parsed. 
For term matches that the LLM returned different label than human annotators, we used the justification text generated by the model to suggest possible causes for the disagreement.

For each round of our prompt iteration, we used 200 tagged terms to test the performance. 
During those tests, GPT-3.5 consistently showed inferior performance compared to GPT-4, so we did not run a full-scale evaluation for GPT-3.5, choosing to focus on GPT-4 only.
For the final full-scale test, we sent 1,500 term matches to GPT-4 (corresponding to 1,500 rows in the tables for human annotators), using our final version of the prompt.

Treating the consensus of two human annotators as the ground truth and grouping all ``Uncertain'' or ``Disagree'' (if the two annotators gave different annotations) labels with the negative label, we find a Matthews correlation coefficient (MCC) of $0.55$ for GPT-4's classification results. This improves slightly to $0.58$ if we discard all ``Uncertain'' or ``Disagree'' from both GPT-4 and human annotators' annotation. 
In particular, GPT-4 tended to label correctly matched terms (``True Positive'' in the annotation guideline and ``match'' for short in the following tables) as incorrectly matched (``mismatch'' for short in the tables). In other words, it is more strict in designating a term as being used in a putatively clincally relevant way.
As shown in \Cref{tbl-chatgpt-master-annotator-2}, for the same set of 1,500 term matches, there are only 29 cases where the master annotator determined it as ``mismatch'' while the annotator 2 thought it was a ``match'', in comparison, there are 285 cases where GPT-4 thought it was ``mismatch'' and the annotator 2 thought it was a ``match''. 

\begin{table}[htbp]
\caption{\label{tbl-chatgpt-master-annotator-2}This table compares GPT-4 annotations to human annotators on 1,500 term matches, illustrating potential bias in GPT-4's classifications. Each term was annotated by GPT-4, the master annotator, and one additional annotator from a pool of human annotators, all following the same previously described guidelines. The concensus between the master annotator and the additional annotator is computed also, with disagreements resulting in an ``uncertain'' label. To prevent confusion with standard confusion matrix terminology, ``true positive'' and ``false positive'' labels have been renamed to ``match'' and ``mismatch'' respectively in this table. The data highlights GPT-4’s tendency to classify terms that `match' as `mismatch' compared to human annotators. For context, the agreement between the annotator pool and the master annotator is also displayed. }
\centering
{\scriptsize
\begin{tabular}{cc|ccc|ccc|ccc}
\cline{3-11} \cline{4-11} \cline{5-11} \cline{6-11} \cline{7-11} \cline{8-11} \cline{9-11} \cline{10-11} \cline{11-11} 
 & \multicolumn{1}{c}{} & \multicolumn{3}{c}{Annotator Pool} & \multicolumn{3}{c}{Master Annotator} & \multicolumn{3}{c}{Concensus}\tabularnewline
 & \multicolumn{1}{c}{} & match & mismatch & uncertain & match & mismatch & uncertain & match & mismatch & uncertain\tabularnewline
\cline{3-11} \cline{4-11} \cline{5-11} \cline{6-11} \cline{7-11} \cline{8-11} \cline{9-11} \cline{10-11} \cline{11-11} 
\multirow{3}{*}{GPT-4} & match & 843 & 34 & 34 & 869 & 20 & 22 & 818 & 10 & 83\tabularnewline
 & mismatch & 285 & 266 & 32 & 290 & 226 & 67 & 220 & 193 & 170\tabularnewline
 & uncertain & 3 & 0 & 3 & 4 & 0 & 2 & 3 & 0 & 3\tabularnewline
\hline 
\end{tabular}

\begin{tabular}{cc|ccc}
 & \multicolumn{1}{c}{} & \multicolumn{3}{c}{Master Annotator}\tabularnewline
 & \multicolumn{1}{c}{} & match & mismatch & uncertain\tabularnewline
\cline{3-5} \cline{4-5} \cline{5-5} 
\multirow{3}{*}{Annotator Pool} & match & 1041 & 29 & 61\tabularnewline
 & mismatch & 75 & 203 & 22\tabularnewline
 & uncertain & 47 & 14 & 8\tabularnewline

\end{tabular}
}
\end{table}

\section{Discussion}
\label{sec:discussion}

In this paper, we show the importance of manual curation in developing a biomedical dictionary to study epilepsy-related discussions on social media.
Despite its cultural importance and global reach, \textit{Instagram} , the social media we study has not been a focus of social media research.
In general, most biomedical research on social media focuses on platforms with freely and easily accessible data, such as \textit{Twitter}  (now named X).
In addition, such studies are either interested in a few drugs or medical-related terms \cite{pierce2017evaluation, ginn2014mining, o2014pharmacovigilance}, or on conditions with less associated stigma.
Very few studies so far have analyzed the social media discourse of people with epilepsy and their caregivers \cite{Wood2022} using biomedical dictionaries.

The use of dictionary-based methods to study biomedical discourse, however, is not new.
In general, they have been used for knowledge discovery and applied to large corpus of clinician-written notes over the course of clinical examination, and more recently extracted from electronic health records \cite{aronsonOverviewMetaMapHistorical2010, huangBiomedicalNamedEntity2020, reateguiComparisonMetaMapCTAKES2018, yangExploitingPerformanceDictionarybased2008}.
As much as these dictionaries have been shown to work well in clinical settings, they have not been tailored to the informal and broad discourse of social media.
Therefore, our goal here using human annotation is to identify and evaluate the impact of false-positive dictionary terms matched to social media data.
We were able to identify and remove terms with high false-positive rates and that were not removed in our initial automated curated process, thus highlighting the importance of a human-in-the-loop throughout the curation process.
Human annotation has been widely used to analyze and understand sentiments, behaviors, perceptions, languages, or relations of people in social media \cite{eryiugit2013turksent, Zubiaga2015}.
Even though automated annotation techniques are able to identify and match terms faster than human annotation \cite{taboada2014automated}, our results support prior literature showing human annotation is necessary when extracting term meaning from social media posts \cite{Cardie2008,aroyo2015truth}.
Therefore, balancing between the speed of automated annotation and the reliability and validity of human annotation can be one of key approaches to optimal results \cite{Cardie2008, Russel2002, Mitchell1997}.

We identified what kinds of terms tend to mean differently than their biomedical meaning through human annotation and how often they have been used in social media posts. 
The false positive matches were from the different meanings of words. Proper nouns were one of the most common cases. 
Besides, words were often used as metaphors, which cannot be learned from the English dictionary. 
These results imply that future research on social media with biomedical dictionaries should be cautious about terms with multiple meanings to perform a more precise analysis. 

Another noise source in term tagging on social media is lack of context information. 
There were cases where our annotators found it difficult to determine the true meaning of the words due to lack of context. 
It would be impossible for machine based automated methods to determine whether those term matches should be tagged or not. 
If there are abundant cases among them that the term was actually not used with the biomedical relevant meaning, they would introduce the same kinds of noise as the ambiguous terms we identified in this study, to the text mining analysis. 

Our results of dictionary refinement via human annotation also demonstrate how a dictionary with ambiguous terms can have a great impact on downstream data analysis. 
We showed that removing a few terms with high false-positive rates from the dictionary, guided by human annotation, could significantly change the results of network analysis. 
We found that the impacts of the refined dictionary were not limited to the terms we removed. 
There is no reason to believe that eigenvector centrality is the only analysis that will benefit from dictionary refinement.
Trivially, network analysis based on node weight generated by node centrality based methods could be affected.  
In addition, in future studies, we can also test the impact of dictionary refinement on other network analysis, including, but not limited to community detection and link prediction, all the way to impact on actual application like information recommendation system. 

The great difference in the impact of the same dictionary refinement on different data sources also suggests potential network-based methods for identifying low-quality terms in the dictionary.
Those low quality terms could be either ambiguous terms we discussed in this study or terms should not be in the dictionary in the first place. 
If our assumption is correct, the majority of the knowledge structure and spurious connections generated by noise shall have distinctively different structure patterns. 
Either seperating the noisy pattern from the majority knowledge structure directly, or simply removing nodes and measuring the impact like we did for the dictionary refinement, could be potentially feasible approaches to identify the low-quality terms.

We also found that some data corpus benefit more from this dictionary refinement than other data corpus, suggesting the need for dictionary refinement should tailor to each individual research topic and data corpus. 
The four additional data sources besides \textit{Instagram} may have different sets of their own ambiguous terms distinct from the 8 identified for \textit{Instagram} . Indeed, the manual annotation focused only on \textit{Instagram} posts and it is possible that a similar approach could be useful for refining dictionaries specific to the EF forums, PubMed abstracts, Clinical Trials and even \textit{Twitter} . 
Due to the associated cost of manual annotation, it is beyond the scope of this work to do so. 
The variation in ambiguous terms across datasets may be attributable to the inherent characteristics and the qualitative distinctions of each data corpus. 
Language usage and richness levels could be different between social media, online health communities, and medical resources. For instance, polysemous words in online health communities and medical resources have much higher chances of being used with their medical-relevant and professional meanings than their casual meanings than in social media. 
Identifying these differences may benefit future researchers in improving the quality of biomedical signal analysis on different types of datasets. 
Especially, it would be essential to be cautious about the noisy dataset, such as data from social media. Using deep learning models, multi-corpus training, and normalization could be other ways to improve the performance of social media mining regarding the differences between those resources \cite{Magge2021.02.09.21251454}.

In \Cref{sec:results}, we have shown that GPT-4, as a representative of the latest LLM, could not replace human annotators in this task. 
There are several factors that contribute to this performance difference. 
The most significant factor we observed was that the GPT-4 often stayed within the narrow definition of some terms in the guidelines. 
For example, in the guidelines, we mentioned that some terms in our dictionary were imported from medical dictionaries like MedDRA. 
Human annotators could understand why ``marriage'' and ``tattoo'' are in MedDRA in the first place, and why ``orange'' as food can be an allergen, then labeled the term matches as ``true positive'' when they thought it was appropriate, while GPT-4 believed that the term was not used as a medical term and therefore should be labeled as ``false positive''. 
A few counterexamples in the prompt can help, but cannot eliminate such kinds of bias. 
GPT-3.5 did much worse than GPT-4 in this kind of bias (see SI \Cref{sec:si:gpt35}). 
For both models, it was hard to enumerate all possible term types for which the model may fail.

In this paper, we discuss the results of using GPT series models as a direct replacement for human annotators. 
However, it is not the purpose of this paper to find the optimal workflow to use LLM on these types of tasks. 
Still, we gained some valuable experience about prompt engineering for the labeling task in this paper and we recommend people to spend considerable time on prompt engineering if they aim to use LLM for similar tasks.
The time spent on prompt engineering should be included into the time cost in project management.
The prompt engineering is not just about writing one good prompt from scratch, but is more about performing a comprehensive test to catch the ``slips'' or the bias of LLM in some types of terms, then trying to correct the bias in many iterations of the prompt. 
This could require substantial amounts of human annotated data, depending on the goal of the research, with more accurate estimation on the term mislabeling need more data, and coarser estimation requires less. 
Some sampling methods could be used to reduce the size of data set needed, but in principle cannot improve the coverage of corner cases. 
If sufficient human annotated data are readily available, fine-tuning the model could potentially be more effective than prompt engineering.
Taking into account all those steps, the use of LLM for term labeling still requires substantial time investment if accuracy is a priority. 

Although current LLMs are not perfect replacements for human annotators, the research on LLM is a fast developing field with better models coming out every year. 
We should not underestimate the potential of LLM in the future application of social media mining. 
Bearing in mind the prerequisites of conducting comprehensive validations prior to the deployment of LLMs, several promising approaches are currently emerging that utilize LLMs to enhance term tagging in the immediate or foreseeable future. 
One possible hybrid approach is discussed above: first uses human annotation to align LLM output, then LLM could be used to annotate larger data corpus. LLM's annotation will then be used for dictionary refinement. In this way, we accumulate much more annotated data, thus providing better coverage for low-frequency terms and leading to potentially better dictionary refinement, with the same amount of human annotation data. 
In the future, LLM could also be used to tag each occurrence of the term directly for the entire text corpus without dictionary refinement, so the ``correct'' usage of high false positive rate terms can be tagged and contribute to downstream analysis, in contrast to discarding all matches of high false-positive rate terms. 
However, even if current LLMs are sufficiently accurate, their high operational costs still prohibit large-scale applications. This limitation is likely to be alleviated in the future.

\section{Conclusion}
This research identified the terms in our dictionaries with higher frequencies and false-positive rates in an epilepsy cohort on \textit{Instagram} by using the human annotation method. 
On comparison, GPT series models cannot replace human annotation in the annotation process. 
We also identified which terms should be removed from our dictionary to obtain more epilepsy-relevant results from network analysis. Additionally, we identify the fundamental cause of incorrect term matches---having different meanings that are often used---and specific cases. Finally, through validation of the impact of the refined dictionary on eigenvector centrality analysis of the co-mention networks, we demonstrated the impacts and complementary role of the human annotation method to the automated annotation technique. 

Our results demonstrate that human-centered dictionary refinement should be performed when conducting social media mining of biomedical relevance, and epilepsy in particular. It remains to be determined if this approach is also necessary for analyzing other types of scientific or patient-centered textual resources.

Our study has utilized a large number of terms in medical dictionaries and focused on \textit{Instagram} , which may have significant potentials for large-scale social media analysis. Other social media studies on medical corpus have more focused on \textit{Twitter}  and \textit{Facebook} , and most of them have selected a few specific medical terms in their analyses. 
Our results and implications may help future research in our research communities, including the areas of bioinformatics and health informatics, by improving analysis of medical topics on social media speeches. 
Ultimately, this research would contribute to developing knowledge graphs that more truthfully represent the underlying knowledge structure, which can be used for personalized information systems that can support PWE, PWEC. And the workflow could also be applied to other diseases and support other disease cohorts. 

\section{Acknowledgement}

This work was partially funded by National Institutes of Health, National Library of Medicine Program, grant 01LM011945-01 (LMR, AM, XW, RBC, and WRM), a Fulbright Commission fellowship (LMR), the NSF-NRT grant 1735095 ``Interdisciplinary Training in Complex Networks and Systems'' (LMR, AM, XW and RBC), and Fundação para a Ciência e a Tecnologia, grant PTDC-MEC-AND-30221-2017 (RBC)
The funders had no role in study design, data collection and analysis, decision to publish, or preparation of the manuscript.

\bibliographystyle{ama}
\bibliography{JAMIA-references}

\appendix

\pagebreak
\addcontentsline{toc}{part}{Supplemental Information} %
\begin{center}
    \Huge
    Supplemental Materials
\end{center}

\setcounter{equation}{0}
\setcounter{table}{0}
\setcounter{figure}{0}
\setcounter{page}{1}
\makeatletter
\renewcommand{\theequation}{S\arabic{equation}}
\renewcommand{\thetable}{S\arabic{table}}
\renewcommand{\thefigure}{S\arabic{figure}}

\begingroup
    \parindent 0pt \leftskip 0cm \rightskip .75cm \parfillskip -\rightskip
    
    \newcommand*{\EndParWithPagenoInMargin}
    {\nobreak\hfill
    \makebox[0.75cm][r]{\mdseries\normalsize\etocpage}%
    \par}
    \renewcommand*\etoctoclineleaders
        {\hbox{\normalfont\normalsize\hbox to .75ex {\hss.\hss}}}
    \newcommand*{\EndParWithPagenoInMarginAndLeaders}
        {\nobreak\leaders\etoctoclineleaders\hfill
        \makebox[0.75cm][r]{\mdseries\normalsize\etocpage}%
        \par }

    \renewcommand{\etoctoprule}{\hrule height 1pt}
    \renewcommand{\etoctoprulecolorcmd}{\color{black!25}}

    \renewcommand\etoctoprule {\hrule height 1pt\relax }
    \renewcommand\etoctoprulecolorcmd {\color{black!25}}
    \renewcommand\etocaftercontentshook
        {\medskip\begingroup \color{black!25}\hrule height 1pt \endgroup }

    \etocruledstyle[1]{}

    \etocsetstyle {section}
    {}
    {\leavevmode\leftskip 1cm\relax}
    {\bfseries\large\llap{\makebox[1cm][r]{\etocnumber\ \ }}%
    \etocname\EndParWithPagenoInMargin\smallskip}
    {}
    \etocsetstyle {subsection}
    {}
    {\leavevmode\leftskip 2.25cm\relax}
    {\mdseries\normalsize\llap{\makebox[1.25cm][l]{\etocnumber}}%
    \etocname\EndParWithPagenoInMarginAndLeaders}
    {}
    \etocsetstyle {subsubsection}
    {}
    {\leavevmode\leftskip 2.75cm\relax }
    {\mdseries\normalsize\llap{\makebox[1cm][l]{\etocnumber}}%
    \etocname\EndParWithPagenoInMarginAndLeaders}
    {}
    
    \localtableofcontents
    \label{toc:si}
\endgroup

\pagebreak

\FloatBarrier
\section{Annotation process \& analysis example}

\begin{figure}
    \centering
    \includegraphics[width=0.9\linewidth]{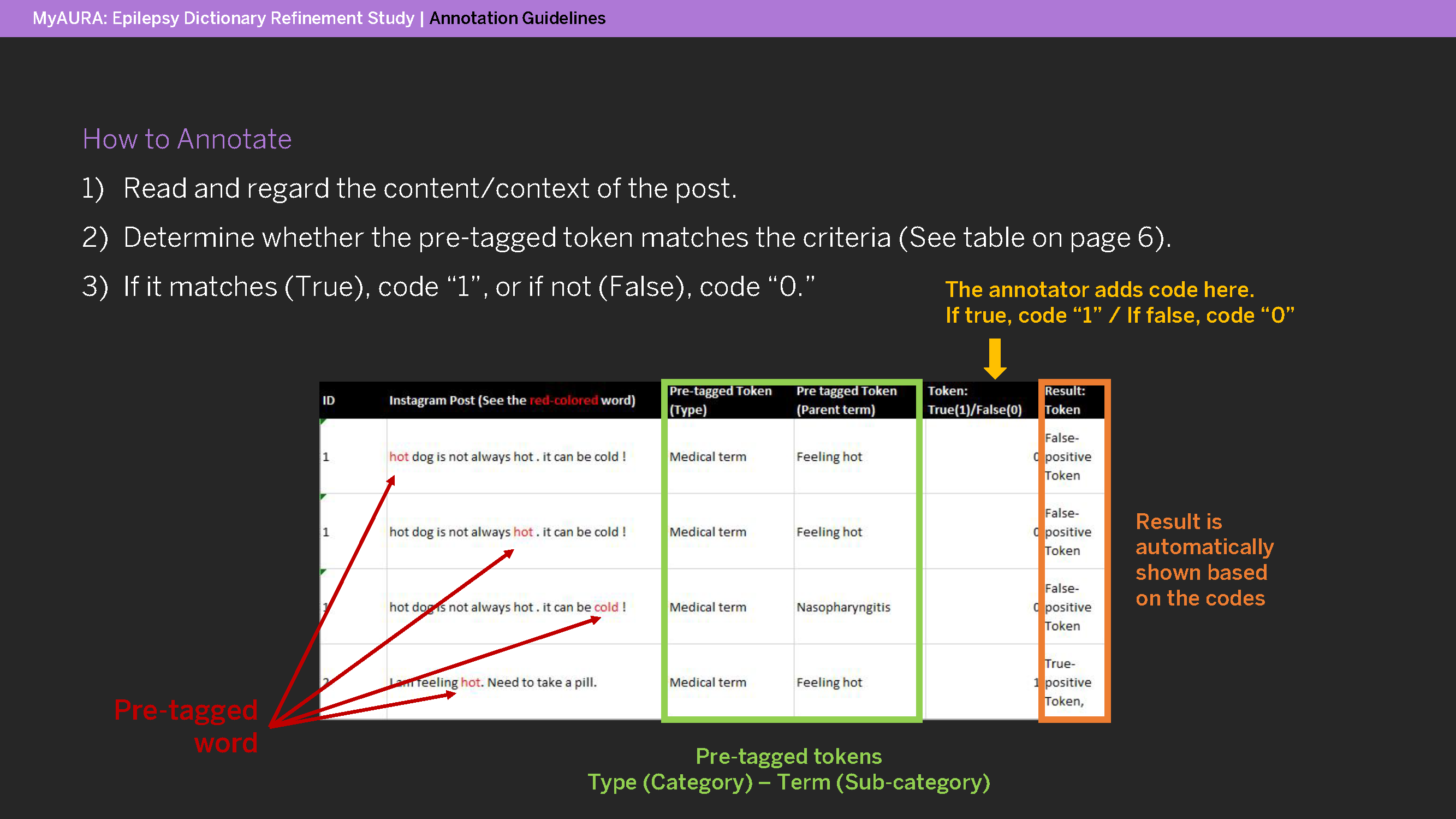}
    \caption{
        Screenshot of annotation guidelines
    }
    \label{si:ant_guide_exmpl}
\end{figure}

\begin{figure}
    \centering
    \includegraphics[width=0.9\linewidth]{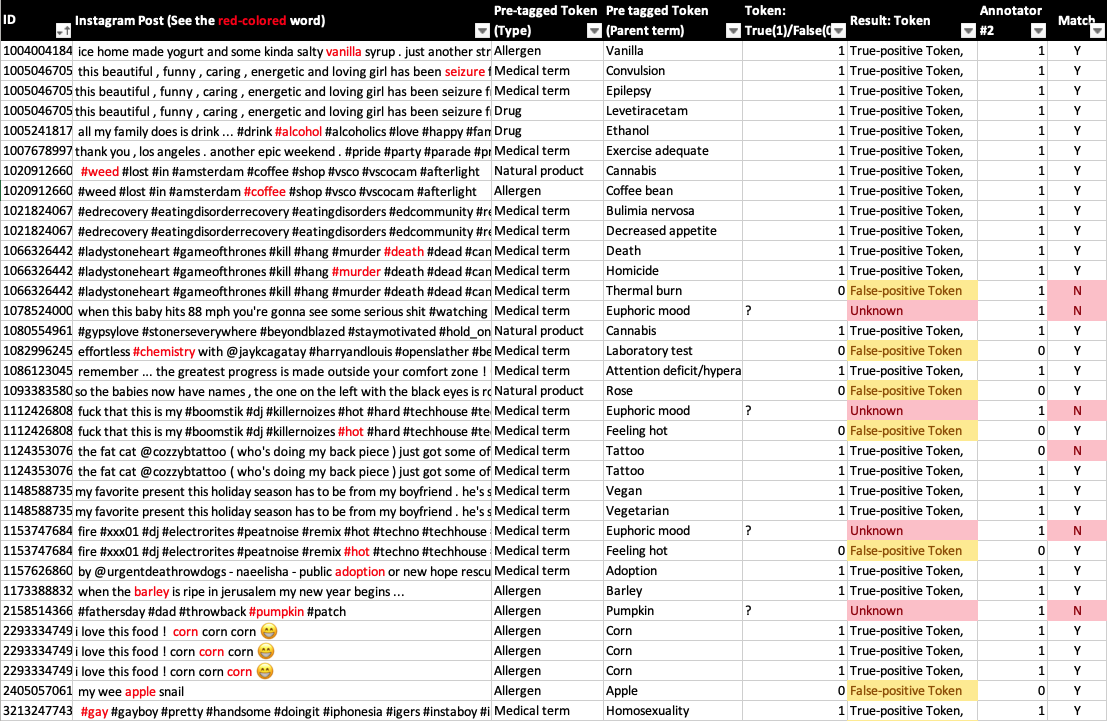}
    \caption{
        Screenshot of annotation analysis
    }
    \label{si:ant_exmpl}
    \label{fig:annotation}
\end{figure}

\FloatBarrier
\section{Dictionary refinement impact on knowledge networks}
\label{sec:si:refinement-effects}

Here we list the tables comparing the top scoring terms for eigenvector centrality on knowledge networks built from both the original and the refined dictionary.
We study the effect of both dictionaries on four different data sources, including:
(i) the epilepsy Instagram cohort (as discussed in main manuscript),
(ii) the Epilepsy Foundation (EF) discussion forums,
(iii) a selection of epilepsy-related abstracts retrieved from PubMed, and
(iv) epilepsy-related Clinical Trials (CT) retrieved from \href{http://clinicaltrials.gov}{clinicaltrials.gov}.
Sections below contain additional details of each data source as well as scoring results.

\subsection{Construction of co-mention network}
\label{si:sec:comention}

We constructed a simple co-mention counts network for each of the data sources. In this network, each node is a parent term that occurs in any of the posts in the dataset. The edge weight is the number of posts containing both term A and term B.

The text unit ``post'' has a different definition for each data source. For social media like Instagram, the Epilepsy Foundation forum and Twitter, a post is a post. For PubMed, the text unit is the title plus the abstract text. For clinical trials, we concatenate all text fields in one clinical trial record into a single ``post''.

The co-mention networks is built on the level of parent terms in the sense that each node is a parent term covering several synonyms in the dictionary. 
The term matching is still done on the level of child terms, in which the dictionary refinement takes place. 
Multiple occurrences of terms under the same parent term will be counted only once. 
Removing a child term does not remove its sibling terms. 
So the term's parent term, representing the cluster of terms, could still be in the co-mention network. 
Removing a term that happens to be the parent term of a cluster of terms also does not exclude its usage as a node label in the co-mention network, but only exclude it from the term matching process.

Due to the above complications, the nodes that disappeared after dictionary refinement are not necessarily the terms we removed.  
Removing a term does not directly remove its parent term from the co-mention network unless it is the only term in the cluster of terms. 
Though removing a child term that counts for the majority of the cluster of terms' occurrence will greatly reduce the co-mention counts and the related node degree. 
Nevertheless, those terms whose all co-mentions are with the removed terms before term removal, will be removed.

As described in our previous research, simple co-mention counts could be a bad choice to study the distance and shortest path between nodes and Jaccard Index is a good normalization method to obtain the proximity network \cite{simas2015distance,Correia2016}. However, for the purpose of obtaining top eigenvector centrality nodes, simple co-mention count is still a good choice while Jaccard Index normalization could potentially suppress too much high frequency terms' weight. In principle, eigenvector centrality on co-mention networks reveal the important nodes by assuming that nodes can vote for each other and important nodes get more votes from other important nodes. If normalization should be in place, it better take place at the level of all edges of a node, not between a pair of nodes.

\subsection{Comparison methods}

Like what we did in the main text, we show the reader two tables, one is the top 20 terms ranking by eigenvector centrality, before and after the dictionary refinement. The other is a table for top 10, 20, 50, 100, 200, and 500 terms lists before and after the dictionary refinement. 
To help the reader get a better idea of if Fagin's generalized Kendall's distance is a big or small distance, we added another column, normalizing the distance with the maximum possible distance for the given $k$ and a given common element numbers between the two lists. 
This is the tricky part of Fagin's generalized Kendall's distance: Kendall's distance is calculated for every pair of items between two lists. The total number of unique elements of two top $k$ lists will not be $k$ but $2k-z$, in which $z$ is the number of common elements between two lists.
Thus it is not advisable to use the normalized distance to compare between different data sources. For that purpose, the raw distance should be used as in \Cref{tab:data-sources-comparison}.

\FloatBarrier
\subsection{Instagram}
\label{sec:si:instagram}

This section presents the knowledge network results for the Instagram data source.
Data details are described in the main text, \cref{sec:methods}.

\begin{table}[htbp]
\caption{\label{tab:instagram-tau-stat}Fagin's generalized Kendall's distance $K$, (missing pair penalty parameter, \(p=0.5\)) for the top \(k\) highest eigenvector centrality terms comparing the Instagram co-mention network built with the original versus the refined dictionary. The third column is the distance normalized by the maximum possible $K$ for a given $k$ and a given count of common elements, so the range of values is $[0,1]$. In the fourth column, we show the average $K$ in the null model described in the main text: 1,000 networks built with 8 random-sampled terms with low FPR and similar frequency. The last column is the standard deviation of $K$ for the null model. }
\centering
\begin{tabular}{rrrrr}
\toprule
$k$ & selected terms $K$ & normalized $K$ & random term $\bar{K}$ & $\sigma$ of random term $K$ \\[0pt]
\midrule
10 & 108 & 0.706 & 48.1 & 16.0\\[0pt]
20 & 393 & 0.701 & 162.1 & 40.5\\[0pt]
50 & 1734 & 0.642 & 613.0 & 84.8\\[0pt]
100 & 6134 & 0.668 & 1528.6 & 381.6\\[0pt]
200 & 19021 & 0.611 & 4420.1 & 1153.6\\[0pt]
500 & 101160 & 0.569 & 27797.9 & 6336.6\\[0pt]
\bottomrule
\end{tabular}
\end{table}

\FloatBarrier
\subsection{Twitter}
\label{sec:si:twitter}

This section presents the knowledge network results for the Twitter data source. 
We parsed Twitter posts from 2006 to 2012 and selected users who have mentioned epilepsy-related drugs or hashtag \#seizuremeds. 
The criteria we used to select Twitter users are the same as Instagram users, which are described in the main text, \cref{sec:methods}.

\begin{table}[htbp]
    \centering
    \footnotesize
    \begin{tabular}{rlclc}
    \toprule
& \multicolumn{2}{c}{Original} & \multicolumn{2}{c}{Refined} \\
\midrule
Rank & Parent term & Eigen-centrality & Parent term & Eigen-centrality\\
\midrule
156 & Pyridoxine & 0.575 & Pyridoxine & 0.577\\[0pt]
1701 & Pantothenic acid & 0.575 & Pantothenic acid & 0.577\\[0pt]
614 & Niacin & 0.574 & Niacin & 0.576\\[0pt]
8595 & Coffee bean & 0.057 & Coffee bean & 0.036\\[0pt]
176626 & Feeling hot & 0.049 & Tea leaf & 0.014\\[0pt]
8976 & Cocoa & 0.035 & Caffeine & 0.009\\[0pt]
8649 & Tea leaf & 0.033 & Somnolence & 0.008\\[0pt]
181534 & Nasopharyngitis & 0.031 & Cocoa & 0.008\\[0pt]
192 & Caffeine & 0.013 & Ethanol & 0.005\\[0pt]
186129 & Somnolence & 0.011 & Riboflavin & 0.004\\[0pt]
815 & Diazepam & 0.010 & Acetaminophen & 0.004\\[0pt]
8592 & Chicken & 0.009 & Cocaine & 0.004\\[0pt]
8645 & Vanilla & 0.007 & Vanilla & 0.004\\[0pt]
884 & Ethanol & 0.007 & Cyanocobalamin & 0.004\\[0pt]
307 & Acetaminophen & 0.007 & Thiamine & 0.004\\[0pt]
8608 & Lemon & 0.006 & Cherry & 0.003\\[0pt]
178464 & Hyperhidrosis & 0.005 & Nicotinamide & 0.003\\[0pt]
8633 & Rice & 0.005 & Cannabis & 0.003\\[0pt]
177790 & Headache & 0.005 & Scratch & 0.003\\[0pt]
8575 & Apple & 0.005 & Goose & 0.003\\[0pt]
    \bottomrule 
    \end{tabular}
    \caption{
        Top 20 eigenvector centrality terms of the epilepsy cohort Twitter knowledge networks built with the original (left) and the refined (right) dictionary.
    }
    \label{tab:si:twitter-eigen}
\end{table}

\begin{table}[htbp]
\caption{\label{tab:twitter-tau}Fagin's generalized Kendall's distance, (missing pair penalty parameter, \(p=0.5\)) for the top \(k\) highest eigenvector centrality terms comparing the Twitter co-mention network built with the original versus the refined dictionary. The last column is the distance normalized by the maximum possible distance so the value range is $[0,1]$. }
\centering
\begin{tabular}{rrr}
\toprule
$k$ & generalized Kendall's distance & normalized distance\\[0pt]
\midrule
10 & 34 & 0.515\\[0pt]
20 & 300 & 0.739\\[0pt]
50 & 974 & 0.515\\[0pt]
100 & 2418 & 0.396\\[0pt]
200 & 8607 & 0.357\\[0pt]
500 & 46050 & 0.318\\[0pt]
\bottomrule
\end{tabular}
\end{table}

\FloatBarrier
\subsection{Epilepsy Foundation forums}
\label{sec:si:eff}

This section presents the knowledge network results for the Epilepsy Foundation (EF) data source.
Data were collected from the forum of \href{http://www.epilepsy.com}{www.epilepsy.com}.
Forums contained long textual descriptions of epilepsy-related discussions separated by individual questions and answers categorized by topics.

\Cref{tab:si:eff-eigen} shows the top eigenvector centrality terms of the knowledge network built from this data source before and after dictionary refinement. 
Note how false-positive term matches did not have a great impact over the top ranking terms.

\begin{table}[htbp]
    \centering
    \footnotesize
    \begin{tabular}{rlclc}
\toprule
& \multicolumn{2}{c}{Original} & \multicolumn{2}{c}{Refined} \\
\midrule
Rank & Parent term & Eigen-centrality & Parent term & Eigen-centrality\\
\midrule
1 & Convulsion & 0.595 & Convulsion & 0.596\\[0pt]
2 & Epilepsy & 0.492 & Epilepsy & 0.493\\[0pt]
3 & Levetiracetam & 0.249 & Levetiracetam & 0.249\\[0pt]
4 & Electroencephalogram & 0.232 & Electroencephalogram & 0.232\\[0pt]
5 & Grand mal convulsion & 0.198 & Grand mal convulsion & 0.199\\[0pt]
6 & Lamotrigine & 0.195 & Lamotrigine & 0.195\\[0pt]
7 & Surgery & 0.140 & Surgery & 0.140\\[0pt]
8 & Anxiety & 0.133 & Anxiety & 0.134\\[0pt]
9 & Valproic Acid & 0.124 & Valproic Acid & 0.124\\[0pt]
10 & Stress & 0.120 & Stress & 0.120\\[0pt]
11 & Phenytoin & 0.098 & Phenytoin & 0.099\\[0pt]
12 & Carbamazepine & 0.096 & Carbamazepine & 0.097\\[0pt]
13 & Topiramate & 0.093 & Topiramate & 0.093\\[0pt]
14 & Confusional state & 0.092 & Confusional state & 0.092\\[0pt]
15 & Depression & 0.083 & Depression & 0.083\\[0pt]
16 & Partial seizures & 0.069 & Partial seizures & 0.069\\[0pt]
17 & Pain & 0.068 & Pain & 0.068\\[0pt]
18 & Pregnancy & 0.068 & Pregnancy & 0.068\\[0pt]
19 & Aura & 0.067 & Aura & 0.067\\[0pt]
20 & Amnesia & 0.066 & Amnesia & 0.066\\[0pt]
\bottomrule 
    \end{tabular}
    \caption{
        Top 20 highest eigenvector centrality terms of the forum of epilepsy.com knowledge networks built with the original (left) and the refined (right) dictionary.
    }
    \label{tab:si:eff-eigen}
\end{table}

\begin{table}[htbp]
\caption{\label{tab:eff-tau}Fagin's generalized Kendall's distance, (missing pair penalty parameter, \(p=0.5\)) for the top \(k\) highest eigenvector centrality terms comparing the epilepsy.com forum co-mention network built with the original versus the refined dictionary. The last column is the distance normalized by the maximum possible distance so the value range is $[0,1]$. }
\centering
\begin{tabular}{rrr}
\toprule
$k$ & generalized Kendall's distance & normalized distance \\[0pt]
\midrule
10 & 0 & 0.000\\[0pt]
20 & 0 & 0.000\\[0pt]
50 & 101 & 0.079\\[0pt]
100 & 344 & 0.067\\[0pt]
200 & 947 & 0.046\\[0pt]
500 & 1099 & 0.009\\[0pt]
\bottomrule
\end{tabular}
\end{table}

\FloatBarrier
\subsection{PubMed}
\label{sec:si-pubmed}

This section presents the knowledge network results for the PubMed (Medline) data source.
We selected papers from PubMed if they satisfy any of the following criteria:
(i) they were published on a expert-curated list of 18 epilepsy-focused journals (see \cref{tab:si:pubmed-journal});
(ii) they contained at least one epilepsy-related MeSH term (see \cref{tab:si:pubmed-mesh}; or
(iii) their titles or abstracts contain the name of at least one drug known to treat epilepsy (as described in \cref{sec:methods} and listed in \cref{tab:si:pubmed-drug}).
Dictionary terms were matched in both the title and the abstract of all selected papers.

\begin{table}
    \centering
    \footnotesize
    \begin{tabular}{rl}
    \toprule
    & Journal name \\
    \midrule
1  & Clinical nursing practice in epilepsy \\
2  & Epilepsia \\
3  & Epilepsia open \\
4  & Epilepsy \& behavior case reports \\
5  & Epilepsy \& behavior : E\&B \\
6  & Epilepsy \& behavior reports \\
7  & Epilepsy currents \\
8  & Epilepsy journal \\
9  & Epilepsy research \\
10 & Epilepsy research and treatment \\
11 & Epilepsy research. Supplement \\
12 & Epileptic disorders: international epilepsy journal with videotape \\
13 & Journal of epilepsy \\
14 & Journal of epilepsy research \\
15 & Journal of pediatric epilepsy \\
16 & Molecular \& cellular epilepsy \\
17 & Newsletter. American Epilepsy League \\
18 & North African and Middle East epilepsy journal \\
    \bottomrule
    \end{tabular}
    \caption{
        List of epilepsy-focused journals used to build the PubMed dataset.
        All titles and abstracts from articles published in these journals were extracted and used to build the PubMed knowledge network.
    }
    \label{tab:si:pubmed-journal}
\end{table}

\begin{table} 
    \centering
    \footnotesize
    \begin{tabular}{rll}
    \toprule
    & MeSH ID & MeSH term\\
    \midrule
1  & D012640 & Seizures\\
2  & D004827 & Epilepsy\\
3  & D000069279 & Drug Resistant Epilepsy\\
4  & D004828 & Epilepsies, Partial\\
5  & D020936 & Epilepsy, Benign Neonatal\\
6  & D004829 & Epilepsy, Generalized\\
7  & D004834 & Epilepsy, Post-Traumatic\\
8  & D020195 & Epilepsy, Reflex\\
9  & D000073376 & Epileptic Syndromes\\
10 & D000080485 & Sudden Unexpected Death in Epilepsy\\
11 & D000078306 & clobazam (onfi)\\
12 & D000077287 & levetiracetam (keppra)\\
13 & D000077213 & lamotrigine\\
14 & D000078334 & lacosamide\\
15 & D002220 & carbamazepine\\
16 & D003975 & diazepam (valium)\\
17 & D000078330 & xcarbazepine\\
    \bottomrule
    \end{tabular}
    \caption{
        List of epilepsy-focused PubMed MeSH terms used to build the PubMed dataset.
        All titles and abstracts from articles containing these MeSH terms were extracted and used to build the PubMed knowledge network.
    }
    \label{tab:si:pubmed-mesh}
\end{table}

\begin{table} 
    \centering
    \footnotesize
    \begin{tabular}{rl}
    \toprule
    & Drug name \\
\midrule
1  & Levetiracetamum \\
2  & Valium \\
3  & Levetiracetam \\
4  & Diazepam \\
5  & Clobazam \\
6  & Keppra \\
7  & Oxcarbazepine \\
8  & SPM927 \\
9  & Carbamazepen \\
10 & Erlosamide \\
11 & Harkoseride \\
12 & Lamotriginum \\
13 & Lamotrigina \\
14 & Carbamazepina \\
15 & Lamotrigine \\
16 & Carbamazepinum \\
17 & Carbamazepine \\
18 & Diastat \\
19 & Carbamazépine \\
20 & Carbamazepin \\
21 & Vimpat \\
22 & Lamictal \\
23 & Lacosamide \\
24 & Onfi \\
    \bottomrule
    \end{tabular}
    \caption{
        List of drug names used to retrieve PubMed articles to build the PubMed dataset.
        All titles and abstracts from articles containing any of these drug names were extracted and used to build the PubMed knowledge network.
    }
    \label{tab:si:pubmed-drug}
\end{table}

Top eigenvector centrality terms can be seen in \Cref{tab:si:pubmed-eigen}
, before and after dictionary refinement.

\begin{table}[htbp]
    \centering
    \footnotesize
    \begin{tabular}{rlclc}
    \toprule
& \multicolumn{2}{c}{Original} & \multicolumn{2}{c}{Refined} \\
\midrule
Rank & Parent term & Eigen-centrality & Parent term & Eigen-centrality\\
\midrule
1 & Convulsion & 0.595 & Convulsion & 0.599\\[0pt]
2 & Epilepsy & 0.562 & Epilepsy & 0.573\\[0pt]
3 & Electroencephalogram & 0.283 & Electroencephalogram & 0.282\\[0pt]
4 & Surgery & 0.166 & Surgery & 0.170\\[0pt]
5 & Partial seizures & 0.158 & Partial seizures & 0.157\\[0pt]
6 & Valproic Acid & 0.116 & Status epilepticus & 0.112\\[0pt]
7 & Carbamazepine & 0.114 & Valproic Acid & 0.109\\[0pt]
8 & Status epilepticus & 0.111 & Hippocampus & 0.103\\[0pt]
9 & Hippocampus & 0.108 & Carbamazepine & 0.102\\[0pt]
10 & Childhood & 0.103 & Childhood & 0.099\\[0pt]
11 & Temporal lobe epilepsy & 0.092 & Temporal lobe epilepsy & 0.093\\[0pt]
12 & Phenytoin & 0.091 & Phenytoin & 0.080\\[0pt]
13 & Diazepam & 0.081 & Agitation & 0.078\\[0pt]
14 & Agitation & 0.078 & Diazepam & 0.070\\[0pt]
15 & Injection & 0.072 & Grand mal convulsion & 0.067\\[0pt]
16 & Phenobarbital & 0.070 & Injection & 0.066\\[0pt]
17 & Grand mal convulsion & 0.068 & Nervousness & 0.065\\[0pt]
18 & Nervousness & 0.067 & Levetiracetam & 0.064\\[0pt]
19 & Death & 0.065 & Depression & 0.064\\[0pt]
20 & Lamotrigine & 0.061 & Phenobarbital & 0.062\\[0pt]
\bottomrule 
    \end{tabular}
    \caption{
        Top 20 highest eigenvector centrality terms of the epilepsy related PubMed abstracts knowledge networks built with the original (left) and the refined (right) dictionary.
    }
    \label{tab:si:pubmed-eigen}
\end{table}

\begin{table}[htbp]
\caption{\label{tab:pubmed-tau}Fagin's generalized Kendall's distance, (missing pair penalty parameter, \(p=0.5\)) for the top \(k\) highest eigenvector centrality terms comparing the PubMed co-mention network built with the original versus the refined dictionary. The last column is the distance normalized by the maximum possible distance so the value range is $[0,1]$. }
\centering
\begin{tabular}{rrr}
\toprule
$k$ & generalized Kendall's distance & normalized distance \\[0pt]
\midrule
10 & 3 & 0.067\\[0pt]
20 & 82 & 0.355\\[0pt]
50 & 300 & 0.218\\[0pt]
100 & 1324 & 0.238\\[0pt]
200 & 3816 & 0.177\\[0pt]
500 & 19594 & 0.147\\[0pt]
\bottomrule
\end{tabular}
\end{table}

\FloatBarrier
\subsection{Clinical Trials}
\label{sec:si-ct}

This section presents the knowledge network results for the Clinical Trials (CT) data source.
All clinical trials were downloaded from \href{https://clinicaltrials.gov/}{clinicaltrials.gov/} and then filtered if they met any of the following criteria:
(i) the CT conditions field include the word ``epilepsy'';
(ii) the CT intervention field included one of the seven drugs known to treat epilepsy: \textit{Clobazam}, \textit{Levetiracetam}, \textit{Lamotrigine}, \textit{Lacosamide}, \textit{Carbamazepine}, \textit{Diazepam}, and \textit{Oxcarbazepine};
(iii) the CT keywords field included the words ``epilepsy'', ``Anti-epileptic drug'', or ``Seizure'';
(iv) the CT description, title, or summary fields included any of the drug names listed in \cref{tab:si:pubmed-drug}.

We matched our dictionary to the full text description of each clinical trial record.

Top eigenvector centrality terms can be seen in \Cref{tab:si:ct-eigen}
, before and after dictionary refinement.

\begin{table}[htbp]
    \centering
    \footnotesize
    \begin{tabular}{rlclc}
\toprule
& \multicolumn{2}{c}{Original} & \multicolumn{2}{c}{Refined} \\
\midrule
Rank & Parent term & Eigen-centrality & Parent term & Eigen-centrality\\
\midrule
1 & Epilepsy & 0.578 & Epilepsy & 0.579\\[0pt]
2 & Convulsion & 0.577 & Convulsion & 0.577\\[0pt]
3 & Electroencephalogram & 0.251 & Electroencephalogram & 0.251\\[0pt]
4 & Partial seizures & 0.197 & Partial seizures & 0.197\\[0pt]
5 & Surgery & 0.163 & Surgery & 0.163\\[0pt]
6 & Levetiracetam & 0.158 & Levetiracetam & 0.158\\[0pt]
7 & Depression & 0.126 & Depression & 0.126\\[0pt]
8 & Valproic Acid & 0.090 & Valproic Acid & 0.090\\[0pt]
9 & Lamotrigine & 0.089 & Lamotrigine & 0.089\\[0pt]
10 & Anxiety & 0.087 & Anxiety & 0.087\\[0pt]
11 & Carbamazepine & 0.086 & Carbamazepine & 0.086\\[0pt]
12 & Investigation & 0.082 & Investigation & 0.082\\[0pt]
13 & Blindness & 0.081 & Blindness & 0.081\\[0pt]
14 & Agitation & 0.079 & Agitation & 0.079\\[0pt]
15 & Childhood & 0.068 & Childhood & 0.068\\[0pt]
16 & Topiramate & 0.065 & Topiramate & 0.065\\[0pt]
17 & Temporal lobe epilepsy & 0.065 & Temporal lobe epilepsy & 0.065\\[0pt]
18 & Weight & 0.062 & Weight & 0.062\\[0pt]
19 & Prophylaxis & 0.060 & Prophylaxis & 0.060\\[0pt]
20 & Injury & 0.058 & Injury & 0.058\\[0pt]
    \bottomrule 
    \end{tabular}
    \caption{
        Top 20 highest eigenvector centrality terms of the epilepsy clinical trials documents knowledge networks built with the original (left) and the refined (right) dictionary.
    }
    \label{tab:si:ct-eigen}
\end{table}

\begin{table}[htbp]
\caption{\label{tab:ct-tau}Fagin's generalized Kendall's distance, (missing pair penalty parameter, \(p=0.5\)) for the top \(k\) highest eigenvector centrality terms comparing the Clinical Trials co-mention network built with the original versus the refined dictionary. The last column is the distance normalized by the maximum possible distance so the value range is $[0,1]$. }
\centering
\begin{tabular}{rrr}
\toprule
$k$ & generalized Kendall's distance & normalized distance \\[0pt]
\midrule
10 & 0 & 0.000\\[0pt]
20 & 0 & 0.000\\[0pt]
50 & 82 & 0.064\\[0pt]
100 & 137 & 0.027\\[0pt]
200 & 270 & 0.013\\[0pt]
500 & 1715 & 0.014\\[0pt]
\bottomrule
\end{tabular}
\end{table}

\subsection{Dictionary refinement impacts on different data sources}

By comparing \Cref{tab:fagin-tau}, \Cref{tab:twitter-tau}, \Cref{tab:eff-tau}
, \Cref{tab:pubmed-tau} and \Cref{tab:ct-tau}, it is clear that the rank difference introduced by dictionary refinement is mostly significant in Instagram dataset, least (nearly no difference) in EFF and Clincal Trials dataset. This demonstrates the need for dictionary refinement for each individual data source. It also shows EFF as a focused social platform for epilepsy, has a high quality data with much less non-epilepsy-related content. 

\section{Comparison with OpenAI's GPT series models on term annotation}

\subsection{the prompt we used for GPT-4}

\begin{lstlisting}[basicstyle=\ttfamily, breaklines=true]
As part of a study aimed at identifying potentially mislabeled terms by an automated term tagging system, you are tasked with classifying social media text excerpts where a term potentially relevant to epilepsy population text analysis has been pre-tagged. We have built a dictionary for this social media text analysis. We have included many medical terms that come from MedDRA and other medical dictionaries. Additionally, as this analysis could potentially uncover hidden correlations between diet, daily habits, and epilepsy conditions, we have also included many terms commonly used in daily English, such as common foods.

The automated tagging system does not recognize multiple meanings of terms. In some texts, a tagged term is used with the meaning for which we added the term to the dictionary; in others, it is not. Your task is to determine whether the tagged term is used with the epilepsy text analysis related meaning for which we included it in the dictionary. To help you discern which meaning is relevant, we will provide you with the "parent_term" and "type" fields in the dictionary.

For each text entry, you will be provided with the following fields:

- `post_text`: The full text of the social media post with the pre-tagged term surrounded by asterisks (e.g., *term*).
- `matched_token`: The specific term that has been pre-tagged and requires classification.
- `parent_term`: A synonym of the tagged term that has the same meaning for the intended meaning or a term whose meaning is closely related to the intended meaning. 
- `type`: The type of term, such as allergen, drug, medical term, or natural product.

Here are a few examples to help you understand: the term "Mary Jane" in our dictionary has a "parent_term" of "Cannabis" and a "type" of "Natural Product". If "Mary Jane" is used in the post to mean Cannabis, then the term is used with the intended meaning and the auto-tagging system has correctly identified this term. However, if "Mary Jane" is used as a person's name in the text, it does not carry the intended meaning of "Cannabis". Another example, the term "Cold", when used in "I got a bad cold", aligns with its parent term "Nasopharyngitis", but in "It is a cold day", it is used with an irrelevant meaning. The term "Orange" has a type of "Allergen", so whenever it is used to describe the citrus fruit (regardless of whether the person is allergic to orange), it is used with the intended meaning. However, the tagging should be considered incorrect if it is used to describe a color.

Based on the information provided and the context of use in the `post_text`, determine whether the `matched_token` is a "True Positive" or a "False Positive" using the criteria below:

1. **True Positive**:
   - The `matched_token` is accurately related to its `parent term` and `type` within the context of the `post_text`.
   - For "Allergen" type, if the term is a food name, as long as the term does mean the food that somebody could be allergic to, no matter whether the post text contains allergic reactions, it is a "True Positive". 

2. **False Positive**:
   - The usage of the `matched_token` does not align with its `type` or `parent_term` in the context of `post_text`.
   - If the term is used metaphorically or as a proper noun, it is most likely to be used with an unrelated meaning.

3. **Uncertain Classification**:
   - If the context or the information provided is insufficient to classify the `matched_token` definitively, classify it as "Uncertain"
   - Explain the reasons for uncertainty and what additional information might assist in classification.

4. **Translation and Cultural Considerations**:
   - If the `post_text` contains language or cultural references that require interpretation, use your multilingual capabilities and cultural knowledge to inform your classification.
   

**Classification Input Example**:

- `post_text`: "I took *Valerian* to help with my epilepsy, and it calmed my nerves."
- `matched_token`: Valerian
- `parent term`: Valerian
- `type`: Drug
- **Classification**: [Your Analysis and Classification Here]

Respond with JSON format with the following schema:

{"token_class": $your_classification, "reason": $your_justification}

$your_classification should be one of "True Positive", "False Positive" or "Uncertain".
You can provide a brief justification for your decision in $your_justification. 
\end{lstlisting}

\subsection{compare GPT-4's annotation with human annotators}

\begin{table}[htbp]
\caption{\label{tbl-gpt-4-consensus}The contingency table between GPT-4's labels and the consensus labels of the two human annotators on the annotation for 1500 tagged terms. In this table, if any of the two annotators give ``uncertain'' answer, or the two annotators did not return the same label, we label the tagged term as ``uncertain''.}
\centering
\begin{tabular}{lrrr}
\toprule
 & human: match & human: mismatch & human: uncertain\\[0pt]
\midrule
GPT-4: match & 818 & 10 & 83\\[0pt]
GPT-4: mismatch & 220 & 193 & 170\\[0pt]
GPT-4: uncertain & 3 & 0 & 3\\[0pt]
\bottomrule
\end{tabular}
\end{table}

\begin{table}[htbp]
\caption{\label{tbl-gpt-4-master}The contingency table between GPT-4's labels and the master annotator's labels on the annotation for 1500 tagged terms. The master annotator is the most experienced annotator.}
\centering
\begin{tabular}{lrrr}
\toprule
 & master: match & master: mismatch & master: uncertain\\[0pt]
\midrule
GPT-4: match & 869 & 20 & 22\\[0pt]
GPT-4: mismatch & 290 & 226 & 67\\[0pt]
GPT-4: uncertain & 4 & 0 & 2\\[0pt]
\bottomrule
\end{tabular}
\end{table}

\begin{table}[htbp]
\caption{\label{tbl-master-annotator-2}The contingency table between master annotator's labels and the other annotator's labels of the two human annotators on the annotation for 1500 tagged terms. Each tagged term was annotated by two annotators and one of them is our master annotator for the data in this table. The annotator 2 in this table is not a sole annotator but several annotators. Therefore this table shows the disgreement between the master annotator and a group of other human annotators.}
\centering
\begin{tabular}{lrrr}
\toprule
 & annotator 2: match & annotator 2: mismatch & annotator 2: uncertain\\[0pt]
\midrule
master: match & 1041 & 75 & 47\\[0pt]
master: mismatch & 29 & 203 & 14\\[0pt]
master: uncertain & 61 & 22 & 8\\[0pt]
\bottomrule
\end{tabular}
\end{table}

\subsection{GPT-3.5's annotation comparison with human annotators}
\label{sec:si:gpt35}

As mentioned in the main text, we only use 200 term matches to evaluate the performance of GPT-3.5. As shown in \Cref{tab:si:gpt35-consensus} and \Cref{tab:si:gpt35-master}, GPT-3.5's performance is significantly worse than that of GPT-4, whose performance is shown in \Cref{tbl-chatgpt-master-annotator-2} and \Cref{tbl-gpt-4-consensus}. 

\begin{table}[htbp]
    \centering
\begin{tabular}{lrrr}
\toprule
 & human : match & human : mismatch & human : uncertain\\[0pt]
\midrule
GPT-3.5: match & 52 & 1 & 2\\[0pt]
GPT-3.5: mismatch & 60 & 42 & 32\\[0pt]
GPT-3.5: uncertain & 7 & 0 & 4\\[0pt]
\bottomrule
\end{tabular}
    \caption{The contingency table between GPT-3.5's labels and the consensus of two humane annotators' labels on the annotation for 1500 tagged terms. }
    \label{tab:si:gpt35-consensus}
\end{table}

\begin{table}[htbp]
    \centering
\begin{tabular}{lrrr}
\toprule
 & master: match & master: mismatch & master: uncertain\\[0pt]
\midrule
GPT-3.5: match & 54 & 1 & 0\\[0pt]
GPT-3.5: mismatch & 70 & 49 & 15\\[0pt]
GPT-3.5: uncertain & 9 & 0 & 2\\[0pt]
\bottomrule
\end{tabular}
    \caption{The contingency table between GPT-3.5's labels and the master annotator's labels on the annotation for 1500 tagged terms. }
    \label{tab:si:gpt35-master}
\end{table}

\end{document}